\documentclass[lettersize,journal]{IEEEtran}
\usepackage{amsmath,amsfonts}
\usepackage{algorithmic}
\usepackage{algorithm}
\usepackage{array}
\usepackage{textcomp}
\usepackage{stfloats}
\usepackage{url}
\usepackage{verbatim}
\usepackage{graphicx}
\usepackage{cite}
\usepackage{booktabs}
\usepackage{xcolor} 
\usepackage{multirow}
\usepackage{graphicx}
\usepackage{subcaption}
\usepackage{enumitem}

\newcommand{\etc}{\textit{etc}}

\newcommand{\br}{\boldsymbol{r}}
\newcommand{\bx}{\boldsymbol{x}}
\newcommand{\bz}{\boldsymbol{z}}

\newcommand{\by}{\boldsymbol{y}}

\newcommand{\bbf}{\boldsymbol{F}}
\newcommand{\bzR}{\boldsymbol{z_R}}

\newcommand{\bepsilon}{\boldsymbol{\epsilon}}

\newcommand{\bW}{\boldsymbol{W}}
\newcommand{\bQ}{\boldsymbol{Q}}
\newcommand{\bK}{\boldsymbol{K}}
\newcommand{\bV}{\boldsymbol{V}}

\newcommand{\bS}{\boldsymbol{S}}

\begin{document}
	
	\title{Double Helix Diffusion for Cross-Domain \\Anomaly Image Generation}

	\author{Linchun~Wu,
		Qin~Zou,~\IEEEmembership{Senior Member,~IEEE},
		Xianbiao~Qi,
		Bo~Du,
		Zhongyuan~Wang,
		Qingquan~Li
		
		
		\thanks{{L.~Wu, Q.~Zou, B.~Du and  Z.~Wang are with the School of Computer Science, Wuhan University, Wuhan 430072, China (E-mails: \{linchun.wu, qzou, dubo, zywang\}@whu.edu.cn).}}
		\thanks{X.~Qi is with Shenzhen Intellifusion Technologies Co Ltd, Shenzhen 518060, China (e-mail: qixianbiao@gmail.com).}
		\thanks{Q.~Li is with the Guangdong Artificial Intelligence and Digital Economy Laboratory (SZ), Shenzhen 518060, China (e-mail: liqq@szu.edu.cn).}
		
	}
	
	\markboth{Journal of \LaTeX\ Class Files, submission, Sep.~2025}%
	{Shell \MakeLowercase{\textit{et al.}}: IEEEtran.cls for IEEE Journals}
	\maketitle
	
	\begin{abstract}
		
		Visual anomaly inspection is critical in manufacturing, yet hampered by the scarcity of real anomaly samples for training robust detectors. Synthetic data generation presents a viable strategy for data augmentation; however, current methods remain constrained by two principal limitations: 1) the generation of anomalies that are structurally inconsistent with the normal background, and 2) the presence of undesirable feature entanglement between synthesized images and their corresponding annotation masks, which undermines the perceptual realism of the output. This paper introduces Double Helix Diffusion (DH-Diff), a novel cross-domain generative framework designed to simultaneously synthesize high-fidelity anomaly images and their pixel-level annotation masks, explicitly addressing these challenges. DH-Diff employs a unique architecture inspired by a double helix, cycling through distinct modules for feature separation, connection, and merging. Specifically, a domain-decoupled attention mechanism mitigates feature entanglement by enhancing image and annotation features independently, and meanwhile a semantic score map alignment module ensures structural authenticity by coherently integrating anomaly foregrounds. DH-Diff offers flexible control via text prompts and optional graphical guidance. Extensive experiments demonstrate that DH-Diff significantly outperforms state-of-the-art methods in diversity and authenticity, leading to significant improvements in downstream anomaly detection performance.
		
	\end{abstract}
	
	\begin{IEEEkeywords}
		Cross-domain generation, Defect detection, Diffusion model, Anomaly generation, Anomaly detection.
	\end{IEEEkeywords}
	
	\section{Introduction}
	
	Visual anomaly detection constitutes a critical and indispensable task across numerous domains, including manufacturing quality control, security surveillance, and infrastructural maintenance~\cite{xing2025recover,jin2025dual,hu2024anomalydiffusion,zhang2023adding}. The objective of this task is to identify and localize irregular patterns or rare deviations from normal conditions, which often correspond to defects, faults, or potential security threats. However, unlike conventional computer vision tasks that rely on large-scale annotated datasets, anomaly detection must often address the inherent challenge of data scarcity: while normal samples are plentiful, anomalous instances are rare, costly to obtain, or even completely absent during the training phase. Such severe data imbalance presents a significant obstacle to developing robust and generalizable anomaly detection systems~\cite{hu2024anomalydiffusion,lin2021few}.

	\begin{figure}[!t]
		\centering
		
		\includegraphics[width=0.9\linewidth]{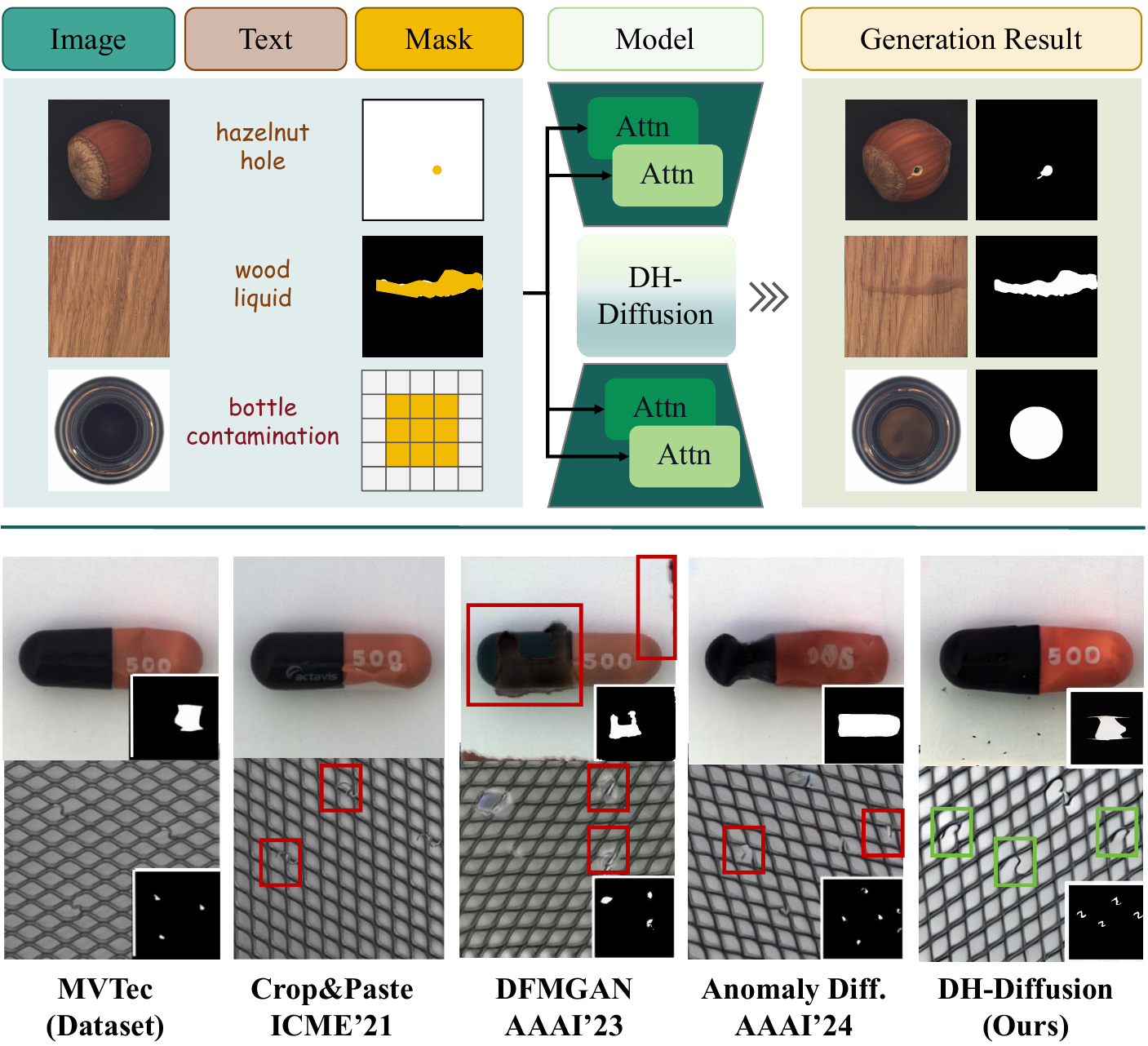}
		
		\caption{The proposed DH-Diff pipeline and some generation results. Top: DH-Diffusion takes the reference image, graphical mask (such as points, detailed masks and raw masks), and text prompts as conditions to generate anomaly images and annotations. Bottom: the anomaly images and masks generated by different methods for the `capsule-squeeze' and `grid-bent' categories on the MVTec dataset.} \label{fig-text1}
	\end{figure}
	
	In the context of data imbalance, a common strategy for anomaly detection involves learning a model of normal pattern using only normal data, particularly in unsupervised settings. Reconstruction-based methods~\cite{xing2025recover,zhu2024toward,liu2022reconstruction,zavrtanik2021reconstruction} are rooted in this idea. These approaches learn to reconstruct normal patterns and subsequently identify anomalies as instances that deviate significantly from their reconstructions. Although they perform effectively on common anomalies, they are trained exclusively on normal data and can exhibit instability or limited generalization when confronted with novel or complex anomaly types that were not represented in the training set.
	
\begin{figure*}[!t]
		\centering
		
		\includegraphics[width=0.8\linewidth]{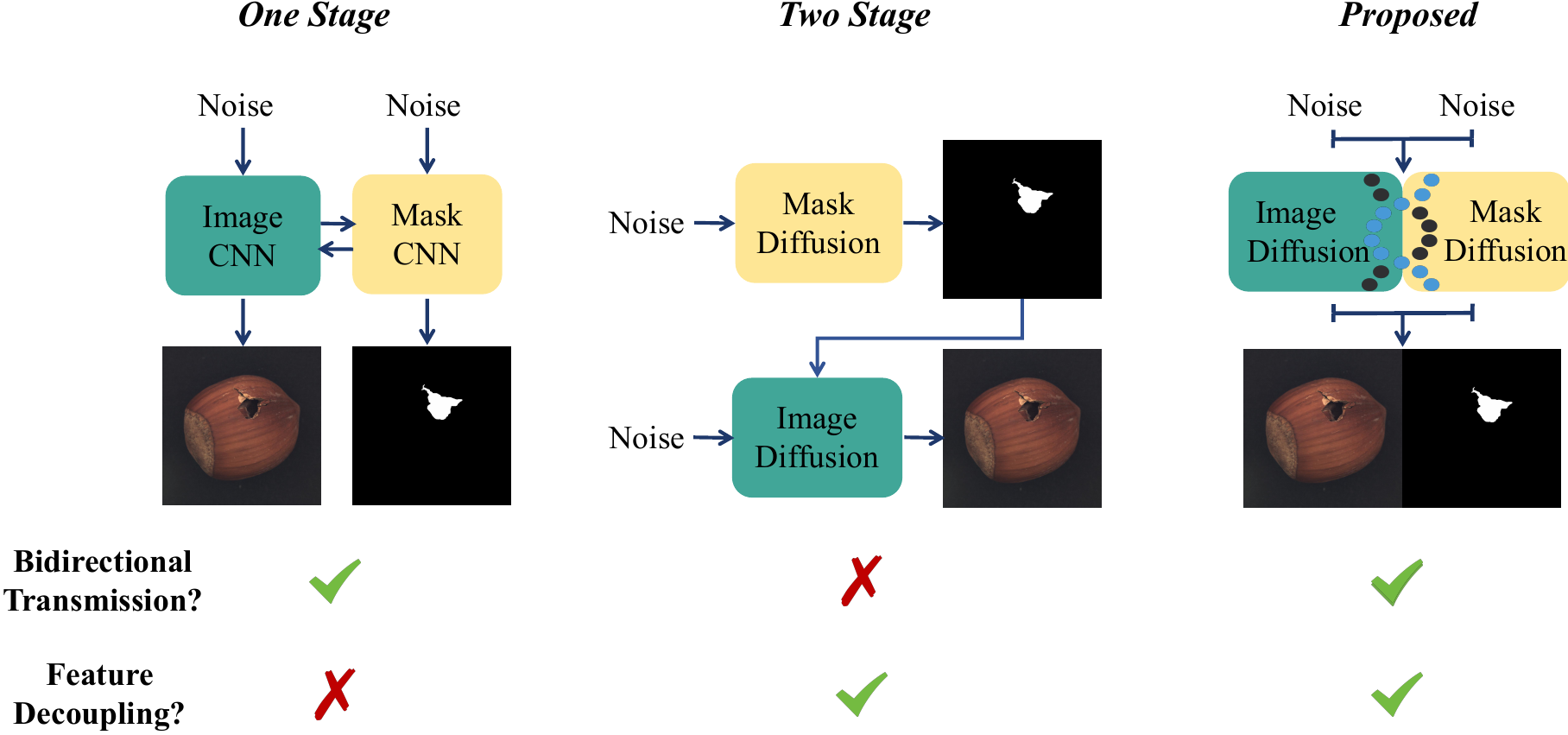}
		
		\caption{Comparison of three anomaly generation frameworks. There are two important properties for generation, i.e., bidirectional transmission and feature decoupling. The former ensures the logicality, while the later guarantees the authenticity. Generally, one-stage methods leverage incorporated image and mask feature for synchronized generation~\cite{duan2023few}, where information are bidirectionally transmitted between the image domain and annotation domain. Two-stage methods first generate a mask, and then perform mask-guided image generation, where features of image and annotation are processed separately. One-stage methods do not guarantee feature decoupling, and Two-stage methods lack bidirectional transmission. The proposed DH-Diff achieves bidirectional transmission while maintaining feature decoupling.}
		\label{fig:text119}
	\end{figure*}

	Another strategy involves synthesizing anomalous samples to enrich the training data, thereby exposing models to a broader spectrum of defects, examples are shown in Figure~\ref{fig-text1}. This approach has emerged as a critical research direction~\cite{jin2025dual,hu2024anomalydiffusion}. Effective data augmentation requires the generation of anomalies that are not only visually diverse but also realistic and structurally consistent with the underlying object or texture. The idea can be implemented in one-stage or two-stage methods, as illustrated in Figure~\ref{fig:text119}. Although existing generation methods have made significant progress in synthesizing anomalies that deviate evidently from normal patterns, they often fall short due to two primary challenges.
	
	The first challenge is structural inconsistency. Existing approaches, particularly two-stage methods that first generate an anomaly mask and then synthesize content within it~\cite{hu2024anomalydiffusion,zhang2024realnet,hua2024image,yang2024slsg}, may yield anomalies that are structurally inconsistent or physically implausible with respect to the normal background. As illustrated in the bottom part of Figure~\ref{fig-text1}, methods such as Anomaly Diff. can sometimes produce artifacts—for instance, a 'bent grid' defect that appears detached from the underlying grid structure, thereby violating inherent object constraints. This issue often arises from the use of predetermined masks that lack semantic coherence with the surrounding context.
	
	The second challenge is feature entanglement. The one-stage methods seek to enhance structural consistency by jointly modeling image and mask information within a single backbone network~\cite{duan2023few}. However, it is easy to generate undesirable artifacts by directly combining or permitting uncontrolled interaction between image and mask features. For instance, image textures may inadvertently 'leak' into the binary annotation mask, or mask features may introduce unnatural biases into the texture of the synthesized anomaly. As illustrated in Figure~\ref{fig:text119}, DFMGAN generates unrealistic brownish content around a capsule.  A fundamental and challenging question remains: can a single backbone efficiently handle the generation of two highly divergent domains — image and annotation?

	\label{sec:related_work}
	
	To address these challenges, a possible solution is to generate both the anomaly image and its precise annotation mask simultaneously, while carefully managing the flow of information between the two domains to ensure both structural integrity and feature purity.
	Inspired by several diffusion editing approaches~\cite{mokady2023null,ma2024subject,jin2025dual}, which suggest that the object structure is mainly defined by the response score map between text prompt and noisy feature, we decide to leverage this characteristic and align the anomalous structure through the score map while maintain the feature purity.
	
	To this end, we propose Double Helix Diffusion, DH-Diff, a novel cross-domain generative framework for anomaly generation in a single backbone. Similar to the structure of double helix, our approach introduces a cyclical interaction between image and annotation features, progressing through dedicated stages of decoupling, connection, and merging within a diffusion model architecture. 
	To relief feature entanglement, we design a Domain-Decoupled Attention (DDA), which  processes image and annotation features through separate attention pathways. 
	To resolve structure conflicts, we introduce Semantic Score Map Modification (SSM), which operates at the semantic level, aligning attention maps derived from text prompts for both the image and annotation. 
	DH-Diff is capable of generating authentic anomaly image-mask pairs simultaneously, offering flexibility through control conditions like text prompts and optional graphical masks. 
	Our main contributions are three-fold:
	\begin{itemize}[leftmargin=*]
		\item A novel cross-domain generative framework, DH-Diff, is proposed to simultaneously synthesize high-fidelity anomaly images and annotation masks, controllable via text and graphical guidance.
		\item A semantic score map modification (SSM) module is introduced to ensure structural logicality and consistency, effectively mitigating structural conflicts observed in previous anomaly synthesis methods.
		\item A domain-decoupled attention (DDA) module is designed to prevent feature entanglement across image and annotation domains, thereby improving generation authenticity.
	\end{itemize}

	The remainder of this paper is organized as follows. Section~\ref{sec:related_work} reviews the related work. Section~\ref{sec:method} details the proposed DH-Diff.  Section~\ref{sec:experiments} presents experiments results. Section~\ref{sec:conclusion} concludes the work.
	
	\section{Related Work}
	\label{sec:related_work}
	
	\subsection{Image Generation}  
	Conventional image generation involves producing an image from Gaussian noise, modeling the transition from a Gaussian distribution to natural images. Early approaches, such as StackGAN \cite{zhang2017stackgan} and TReCS \cite{koh2021text}, laid the groundwork for this field. The advent of large-scale datasets like LAION-5B \cite{schuhmann2022laion} and advancements in diffusion modeling have significantly enhanced text-to-image generation capabilities. Notable models like DALL-E \cite{saharia2022photorealistic} leverage transformer architectures trained on quantized latent spaces. Contemporary state-of-the-art models, including GLIDE \cite{nichol2021glide}, Latent Diffusion Model (LDM) \cite{rombach2022high}, DALL-E-2 \cite{ramesh2022hierarchical}, and Imagen \cite{saharia2022photorealistic}, further push the boundaries of this technology.
	
	With the rapid development of text-to-image models~\cite{nichol2021glide,ramesh2022hierarchical,rombach2022high,saharia2022photorealistic}, image generation has expanded with various type of control signals such as object positions, layouts, scene depth maps, human poses and boundary lines. Models like GLIGEN \cite{li2023gligen} facilitate object layout control, while Make-a-Scene \cite{avrahami2023spatext}, SpaText \cite{avrahami2023spatext}, and ControlNet \cite{zhang2023adding} enable fine-grained spatial control by incorporating semantic segmentation masks into large pre-trained diffusion models. Further advancements include MultiDiffusion \cite{bar2023multidiffusion}, Attend-and-Excite \cite{chefer2023attend}, ReCo \cite{yang2023reco},  and MIGC \cite{zhou2024migc}, which add location controls without fine-tuning the pre-trained text-to-image models.
	
	\subsection{Anomaly Generation}
	
	Anomaly generation, unlike conventional image generation that relies on vast amounts of training data, aims to model anomaly patterns with limited anomaly samples. Current methods can be categorized into two types: 1) random anomaly synthesis and 2) pattern-reliant anomaly modeling.
	
	Random anomaly synthesis involves multiplying an anomaly mask with random textures and applying these augmented textures to normal images. DRAEM \cite{wang2023multimodal} and RealNet \cite{zhang2024realnet} generate abundant anomaly images without relying anomaly training data. However, their results can appear unrealistic, limiting their use for anomaly classification.
	
	Pattern-reliant modeling, on the other hand, learns or copies real anomaly distributions and applies these anomalies to normal images. Approaches such as Cut$\&$Paste \cite{li2021cutpaste}, Crop$\&$Paste \cite{lin2021few}, and PRN \cite{zhang2023prototypical} involve cutting and pasting existing anomalies into normal samples, which may result in limited diversity. Learning-based methods model class-specific anomaly patterns and generate new anomalies on normal samples. SDGAN \cite{niu2020defect} and Defect-GAN \cite{zhang2021defect} generate anomalies from anomalous data but require substantial data and cannot produce anomaly masks. DFMGAN \cite{duan2023few} pre-trains StyleGAN2 \cite{karras2020analyzing} on normal samples and transfers it to the anomaly domain, but it lacks authenticity in generated anomalies and precise alignment between anomalies and masks. Anomaly Diffusion \cite{hu2024anomalydiffusion} learns a spatial anomaly embedding that serves as a text condition to guide anomaly generation.
	
	While these methods are effective for image-level anomaly detection, more sophisticated tasks like pixel-level anomaly segmentation or anomaly classification require pattern-reliant anomaly images with corresponding annotations. Previous methods can be further classified into one-stage and two-stage generation processes. One-stage methods like DFMGAN \cite{duan2023few} generate anomalies with a single backbone, entangle feature from the image and annotation domains, leading to less authentic results. Two-stage methods \cite{hu2024anomalydiffusion,li2021cutpaste,zhang2024realnet} first obtain a mask and then generate or paste anomalies within the masked region of a normal image. However, pre-defined masks may cause structural conflicts with the normal image, resulting in unrealistic anomalies.
	
	In contrast, our model introduces a double-helix structure to relieve the problem of feature entanglement and structure conflicts. This enables the generation of rich, diverse, and authentic anomaly image-mask pairs, enhancing performance in downstream anomaly inspection tasks.

	\subsection{ Anomaly Detection}
	Unlike conventional computer vision tasks \cite{liu2019multistage,jiang2021layercam,zou2018deepcrack} that are trained on millions of labeled data, anomaly detection algorithms typically rely on abundant normal data and sometimes limited anomaly data. With this premises, current anomaly detection methods primarily fall under self-supervised or reconstruction-based. 
	
	Self-supervised approaches dedicated in optimizing the normal decision boundary. Some methods\cite{bergman2020deep,sabokrou2018deep,ruff2018deep,yi2020patch} leverage distribution functions and assume that normal samples conform to a certain distribution function in a high-dimensional space. Various distribution functions are utilized, including Gaussian distribution parameters of normal image embedding vectors \cite{defard2021padim}. Some \cite{ju2015image} do not explicitly calculate distribution functions but instead employ networks to directly find classification boundaries. Examples of network paradigms include few-shot feature residual learning \cite{zhu2024toward}, image transformation prediction \cite{golan2018deep,bergman2020classification} and contrastive learning \cite{tack2020csi}.

	Reconstruction-based algorithms learn the normal structure of samples and reconstruct all test samples, both normal and anomalous, to their normal versions. The anomaly score is then derived from the difference between the reconstructed and input samples.
	Ganomaly~\cite{akcay2019ganomaly} introduces GANs into the reconstruction framework, enhancing the realism of reconstructions through adversarial training. SSPCAB \cite{ristea2022self} and Dream \cite{zavrtanik2021draem} utilize a discriminative end-to-end trainable paradigm for anomaly detection and localization. DiffusionAD \cite{zhang2023diffusionad} incorporates a one-step denoising diffusion process, balancing real-time performance during image reconstruction. DDAD\cite{mousakhan2023anomaly} introduces a multi-step diffusion process, significantly improving image clarity and structural consistency through iterative denoising. RealNet\cite{zhang2024realnet} learns from a pseudo anomaly dataset that includes both anomaly samples and corresponding normal samples, significantly improving reconstruction accuracy.

	Most detection methods are trained using normal samples, but test samples include both normal and anomalous ones. This discrepancy can lead to unpredictable network outputs and hinder performance.
	
	\textbf{Remark.} \emph{The introduced DH-Diff intersects with the three directions mentioned above.} It leverages an image-stable diffusion model to simultaneously generate anomaly image-annotation pairs. The model's ability to produce authentic, structurally coherent anomalies makes DH-Diff highly applicable to various downstream tasks, including anomaly detection and segmentation.

	\begin{figure*}[!t]
		\centering
		\includegraphics[width=1\linewidth]{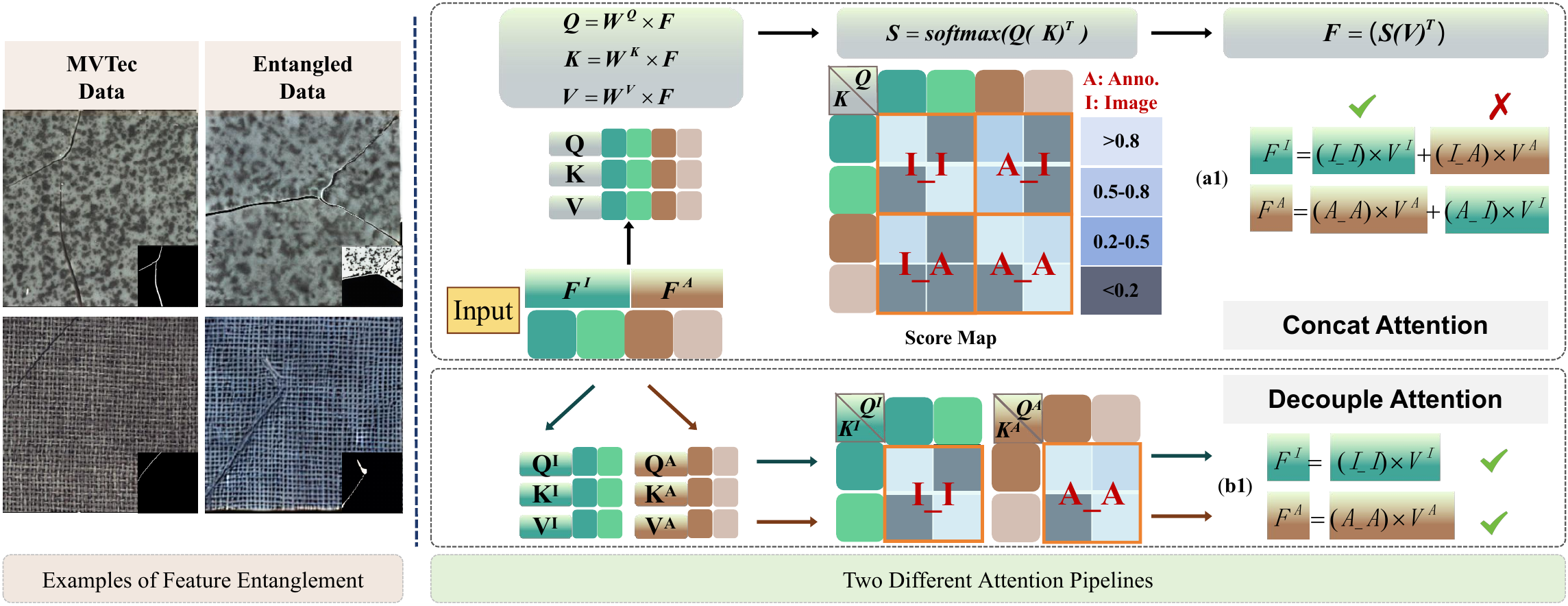}
		\caption{Illustration of the feature entanglement phenomenon (in Left) and the reason causing it (in Right). On the left side, the first column shows two samples from the dataset with perfect textures and masks; the second column shows the generated samples that have drawbacks of mask (for the upper one) and texture color (for the lower one), which are caused  by feature entanglement.
			On the right side, the reason for feature entanglement is illustrated by examining the attention pipelines. In the concatenate attention (top), tokens of image and annotated mask are concatenated as input. When computing the $\text{Score Map}$ of similarity, the structurally similar patches would yield high response scores, which could be found in all four sections, i.e., I\_I (image to image), A\_I (annotation to image), I\_A (image to annotation) and A\_A (annotation to annotation).  Consequently, annotation features $V^{A}$ leak into the image outputs $F^{I}$, and image features $V^{I}$ leak into the annotation outputs $F^{A}$, leading to entangled representations in Eq.~(a1). In contrast, DH-Diff introduces a decouple attention (bottom), which explicitly separates the image and annotation  tokens and computes score map independently by Eq.~(b1),  avoid the feature-entanglement problem.
		}
		\label{fig:entangle}
	\end{figure*}
	
	\section{Method}
	\label{sec:method}
	
	\subsection{Problem Formulation}

	The objective of this work is to achieve joint generation of anomaly images and pixel-wise corresponding masks using a single backbone network, while effectively mitigating feature entanglement and structural conflicts. Our analysis identifies a key issue: direct concatenation of image and annotation features, followed by global feature attention processing, results in cross-domain information entanglement. This phenomenon is reflected in undesirable artifacts in the synthesized images, including distortions in color, structure, and texture, as well as texture leakage in the predicted anomaly masks, as shown in Figure~\ref{fig:entangle}. We argue that these limitations arise primarily from an unconstrained network architecture and an intertwined data flow.

	The unconstrained network architecture component, i.e., the concatenation attention module, is formulated as \( \text{Attention} = \text{softmax}\left(\frac{\bQ \bK^{\top}}{\sqrt{d}} \right) \bV \).
	As illustrated in Figure~\ref{fig:entangle}, when applied to concatenated image–annotation inputs $\bbf = (\bbf^I, \bbf^A)$, the receptive field of $\bQ, \bK$ and $\bV$ simultaneously covers both modalities through:
	\begin{equation}
		\left\{
		\begin{aligned}
			& \bQ = \bW^Q \times (\bbf^I, \bbf^A) = (\bQ_I,\bQ_A),\\
			&  \bK = \bW^K \times (\bbf^I, \bbf^A) = (\bK_I,\bK_A),\\
			&  \bV = \bW^V \times (\bbf^I, \bbf^A) = (\bV_I,\bV_A).
		\end{aligned}
		\right.
		\label{eq:soft}
	\end{equation}  
	
	Consequently, the attention score map
	\(\bS = \text{softmax}(\bQ \bK^{\top})\) produces strong responses between structurally similar patterns across all regions including image–image (I\_I), image–annotation (I\_A), annotation–image (A\_I), and annotation–annotation (A\_A), which can be formulated as:
	\begin{equation}
		\left\{
		\begin{aligned}
			& I\_I = \text{softmax}(\bQ_I\bK_I), I\_A = \text{softmax}(\bQ_I\bK_A),\\
			&  A\_A = \text{softmax}(\bQ_A\bK_A), A\_I = \text{softmax}(\bQ_A\bK_I). \\
		\end{aligned}
		\right.
		\label{eq:soft}
	\end{equation}  
	
	The resulting output feature
	\(\bbf = \bS \bV\) inevitably causes feature leakage. Specifically, the output image feature absorbs annotation information 
	while the output annotation feature absorbs image information through:
	\begin{equation}
		\left\{
		\begin{aligned}
			& \bbf_I = (I\_I) \bV_I + (I\_A) \bV_A,\\
			&  \bbf_A = (A\_A) \bV_A + (A\_I) \bV_I.
		\end{aligned}
		\right.
		\label{eq:soft}
	\end{equation}  
	
	To tackle with this feature entanglement, a possible solution is to design a backbone architecture preserve separate feature flows for image and annotation domains, while still supporting essential cross-domain interactions to maintain structural alignment of both anomalous regions and background content. Recent advances in text-to-image diffusion models~\cite{mokady2023null,ma2024subject} demonstrate the central role of semantic attention in guiding structural formation, where objects foreground is determined by high-response areas between textual prompts and latent features. Inspired by this, we propose the \emph{Semantic Score Map Modification} (SSM) module to achieve cross-domain alignment through semantic attention. Crucially, SSM prevents feature leakage by preserving clean, domain-specific receptive fields for all \(\bQ\), \(\bK\), and \(\bV\), thereby ensuring feature purity across both image–semantic and annotation–semantic pathways.

	To highlight our core innovation, we briefly compare with \textit{DualAno}~\cite{jin2025dual}, the first dual-branch framework for simultaneous image and annotation generation. While \textit{DualAno} separates data flows with parallel backbones and aligns structure using concatenation attention followed by ratio-reduced feature merging and background-based rectification, our approach provides two key advantages: (i) a single-backbone design that simultaneously generates image and annotation features while maintaining domain purity through decoupled attention, and (ii) structural alignment guided by high-level semantic responses rather than low-level feature concatenation. Together, these properties enable efficient joint generation with both strict feature disentanglement and robust structural consistency.

	\begin{figure*}[!t]
		\centering
		\includegraphics[width=0.97\linewidth]{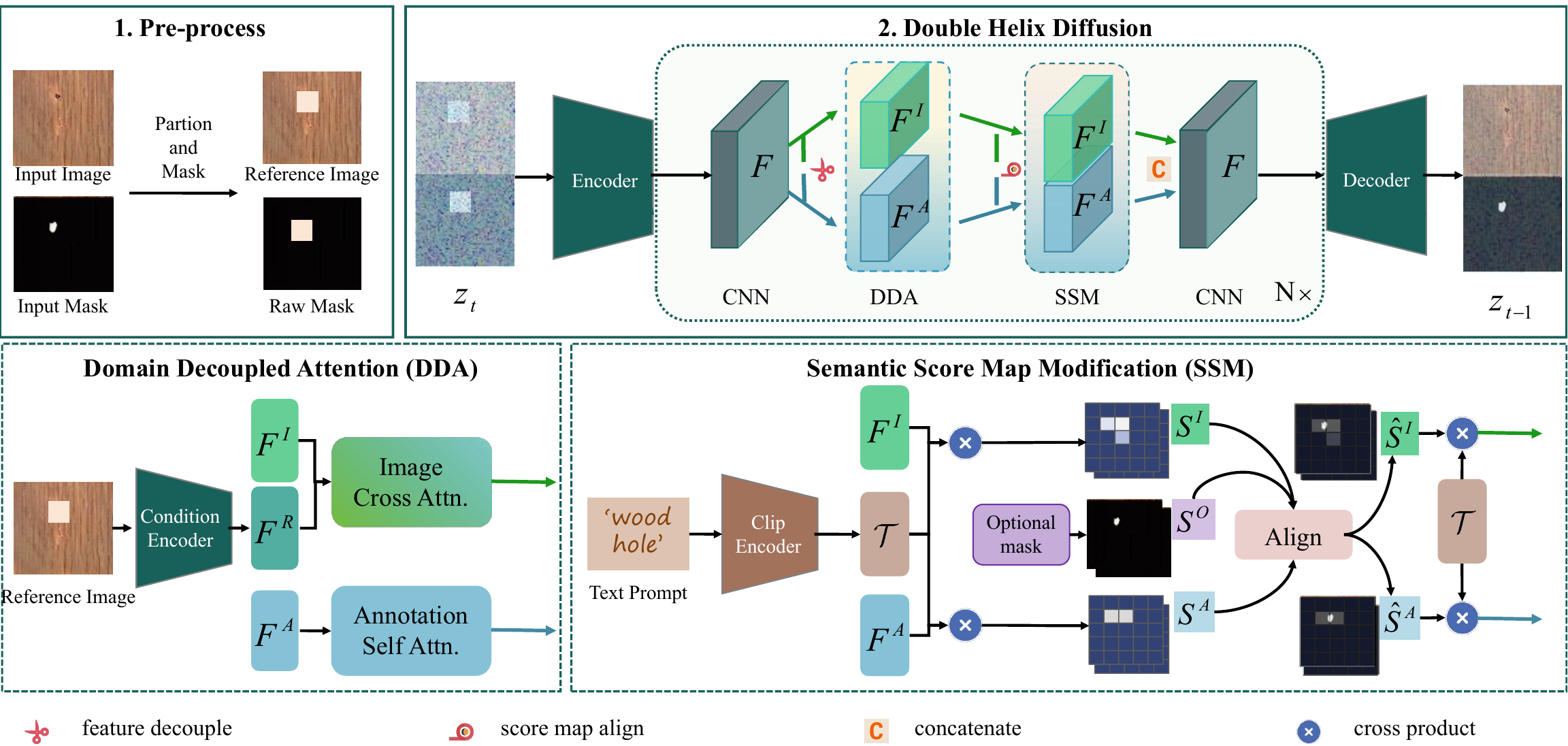}
		\caption{An overview of the proposed DH-Diff. In the pre-processing, given an input anomaly image and its annotation, we calculate the raw mask by dividing the image into $5 \times 5$ patches, and take the regions of the anomaly patches as raw mask. Then, we crop the anomaly image according to the raw mask as reference image. After that, we add noise as the diffusion training/generation input $\bz_t$. In the diffusion process, DH-Diff takes the joint noisy latent $\bz_t$ as input and predicts the noise at time $t$. Subsequently, it deducts the noisy latent $\bz_{t-1}$. The initial features are first processed by domain decoupled attention module (DDA) and then by the semantic score map modification module (SSM). In DDA, we reconstruct the image feature $\bbf^I$ and annotation feature $\bbf^A$ separately, where the reference image feature $\bbf^R$ is incorporated into $\bbf^I$ as  condition. In SSM, we align the score maps from the image domain $\bS_I$ and annotation domain $\bS_A$ with an optional mask $\bS_O$, where the score map is response from noisy feature to text prompt $\mathcal{T}$ and $\bS_O$ is intended for anomaly structure customization.
		}
		\label{fig:framework}
	\end{figure*}
	\subsection{Overview of Our Method}

	Similar to the structure of a double helix, we design our network as a cyclic domain-interaction flow that progresses through three key stages: \textit{decoupling}, \textit{connection}, and \textit{merging}. The decoupling module, i.e., domain decoupled attention (DDA), is tailored to address the problem of feature entanglement, enhances image and annotation representations independently using two parallel attention mechanisms. The connection module, i.e., semantic score map modification (SSM), is devised to mitigate structural conflicts, aligns semantic layouts across the image and annotation domains, thereby ensuring logical consistency and semantic coherence. Furthermore, we provide an optional mask input condition to enable customized control. The interaction trajectory between image and annotation features follows a helix-like projection, hierarchically evolving from coarse to fine scales. This progressive refinement makes the framework highly adaptable to complex visual patterns while maintaining structural integrity. In particular, the cyclic nature of the interaction ensures that complementary cues from both domains are repeatedly reinforced, leading to more stable and coherent anomaly synthesis. Based on this interaction trajectory, we name our framework as DH-Diff -- Double Helix Diffusion.

	As shown in the top of Figure~\ref{fig:framework}, given an input image--mask pair, we first extract a \textit{raw mask} by dividing the image into grids of size \(\tfrac{\text{width}}{K} \times \tfrac{\text{height}}{K}\),  set \(K\)=5 empirically, and label grids that contain anomalies. A \textit{reference image} is then obtained by cropping the regions indicated by the raw mask. Subsequently, we formulate a concatenated input \(\bz = (\bz^I, \bz^A)\), where \(\bz^A\) denotes the noised raw mask feature and \(\bz^I\) the noised reference image feature. This input undergoes iterative processing through the circular CNN, DDA, SSM, and CNN modules over \(N\) cycles, with the encoder comprising the first \(N/2\) cycles and the decoder the remaining \(N/2\). This cyclic pipeline allows information to be exchanged across domains multiple times, progressively enhancing both structural fidelity and semantic alignment.  Details of the DDA and SSM modules are elaborated in the subsequent sections.

	\subsection{Domain Decoupled Attention}  
	
	The input to this module consists of the decoupled image feature $\bbf^I$ and annotation feature $\bbf^A$, as illustrated in the bottom left of Figure~\ref{fig:framework}. Both features share the same tensor shape $[b, c, h \times w]$, where $b$ denotes the batch size, $c$ is the channel dimension, and $h,w$ denote the spatial resolution.  These two features are processed independently through domain-specific attention mechanisms, allowing each branch to capture its own semantic and structural properties before later interaction.  
	
	\textit{Image Cross-Attention.}  
	This module is designed to enhance image texture, color, and structural details by leveraging conditional guidance from a reference feature $\bbf^R$, extracted via a Siamese conditional encoder $E$. To maintain consistency across the forward process, the same diffusion noise $\bepsilon$ applied to $\bbf^I$ is also injected into $\bbf^R$, thereby ensuring that both features share an aligned noise distribution during multi-scale processing. With $\bbf^R$ and $\bbf^I$ as inputs, the attention operation is defined as  
	$\bbf^I = \text{softmax}\!\left(\frac{\bQ^I {\bK^I}^{\top}}{\sqrt{d}} \right)\bV^I,$where the query, key, and value are projected as  
	\begin{equation}
		\left\{
		\begin{aligned}
			\bQ^I &= \bW_Q^{I} \cdot \bbf^I, \\
			\bK^I &= \bW_K^{I} \cdot \big(\zeta_{\omega}(\bbf^R) + \bbf^I \big), \\
			\bV^I &= \bW_V^{I} \cdot \big(\zeta_{\omega}(\bbf^R) + \bbf^I \big).
		\end{aligned}
		\right.
		\label{eq:image_att}
	\end{equation}  
	
	\noindent Here, $\bbf^R = \tau_{\theta}(\bzR)$ denotes the encoded reference feature, with $\tau_{\theta}(\cdot)$ representing the conditional encoder parameterized by $\theta$. The operator $\zeta_{\omega}(\cdot)$ denotes a zero-initialized convolution~\cite{zhang2023adding}, which stabilizes optimization by gradually learning from an initial zero state. $\bW_Q^{I}, \bW_K^{I}, \bW_V^{I}$ are learnable projection matrices. This cross-attention formulation not only preserves the intrinsic structural characteristics of $\bbf^I$ but also enriches them with contextual texture and color information from $\bbf^R$. 
	
	\textit{Annotation Self-Attention.}  
	To transform the coarse input mask into a refined and irregular anomaly mask, we adopt a self-attention mechanism that learns to capture detailed spatial dependencies within the annotation domain:  
	$\bbf^A = \text{softmax}\!\left(\frac{\bQ^A {\bK^A}^{\top}}{\sqrt{d}}\right)\bV^A,$  
	with projections defined as  
	\begin{equation}
		\left\{
		\begin{aligned}
			\bQ^A &= \bW_Q^{A} \cdot \bbf^A, \\
			\bK^A &= \bW_K^{A} \cdot \bbf^A, \\
			\bV^A &= \bW_V^{A} \cdot \bbf^A,
		\end{aligned}
		\right.
		\label{eq:annotation_att}
	\end{equation}  
	
	\noindent where $\bbf^A$ is the flattened annotation feature of shape $[b, c, h \times w]$, and $\bW_Q^{A}, \bW_K^{A}, \bW_V^{A}$ are learnable projection matrices. This mechanism enables the annotation stream to perform fine-grained mask refinement, progressively converting raw binary grids into irregular yet semantically coherent anomaly regions. Importantly, the self-attention operates entirely within the annotation domain, ensuring that spatial priors are faithfully preserved while avoiding unnecessary influence from the image domain. Together, the two attention branches form the foundation of domain decoupling, enabling modality-specific enhancement.

	\subsection{Semantic Score Map Modification}  
	
	The structure of an object, is largely determined by the semantic interaction between the noisy latent feature and the corresponding text embedding \(\mathcal{T}\)~\cite{mokady2023null,ma2024subject}. In particular, cross-attention between the noisy latent and text embedding produces an intermediate semantic score map \( \bS = \bQ \bK^{\top} \), where \(\bQ\) is derived from the latent feature and \(\bK\) from the text embedding. The activated regions within this map indicate the spatial location and shape of the text-related object.  
	
	Motivated by this observation, we propose a cross-domain score map alignment strategy to achieve structurally consistent anomaly generation. The semantic score map from the \textit{image domain} captures realistic and semantically coherent structures, while the map from the \textit{annotation domain} encodes spatial priors derived from the input raw mask. Aligning these two maps allows us to preserve annotation-driven spatial guidance while inheriting structural fidelity from the image, thereby enforcing both semantic authenticity and structural consistency.  
	
	As shown in the bottom-right of Figure~\ref{fig:framework}, SSM module takes the image latent feature \( \bbf^I \) and annotation latent feature \( \bbf^A \) as input. The semantic layouts are extracted as follows:

	\begin{equation}
		\left\{
		\begin{aligned}
			& \bS^I = \text{softmax}(\bQ^I (\bK^{\mathcal{T}})^{\top}),\\
			& \bS^A = \text{softmax}(\bQ^A (\bK^{\mathcal{T}})^{\top}),
		\end{aligned}
		\right.
		\label{eq:soft}
	\end{equation}  
	where \(\bK^{\mathcal{T}}\) is projected from the text condition \(\mathcal{T}\), and \(\bQ^I\), \(\bQ^A\) are queries derived from \( \bbf^I \) and \( \bbf^A \), respectively.  
	
	To further enhance alignment and controllability, we introduce an optional external control mask score map \( \bS^O \). The three maps are fused via a learnable convolutional layer \(\eta\) and a mean function \(\mu\), producing refined attention weights \(\hat{\bS}^I\) and \(\hat{\bS}^A\). These aligned weights guide semantic generation with the text embedding \(\mathcal{T}\), yielding the joint feature \(\bbf\) as:  
	\begin{equation}
		\left\{
		\begin{aligned}
			& (\hat{\bS}^I, \hat{\bS}^A) = \eta(\bS^I, \bS^A, \bS^O) + \mu(\bS^I, \bS^A, \bS^O), \\[4pt]
			& \bbf = C(\hat{\bS}^I \times \bV^{\mathcal{T}}, \; \hat{\bS}^A \times \bV^{\mathcal{T}}),
		\end{aligned}
		\right.
		\label{eq:align_att}
	\end{equation}  
	
	\noindent where \(C(\cdot)\) denotes concatenation, \(\bV^{\mathcal{T}} = \bW_V^{\mathcal{T}} \cdot \mathcal{T}\) is the value projection from the text embedding, and \(\times\) denotes matrix multiplication. This design integrates semantically faithful anomaly structure from the image domain and spatial intent from the annotation domain, while the optional control map \(\bS^O\) provides additional flexibility for customized generation.

	\begin{figure*}[!t]
		\centering
		
		\includegraphics[width=1\linewidth]{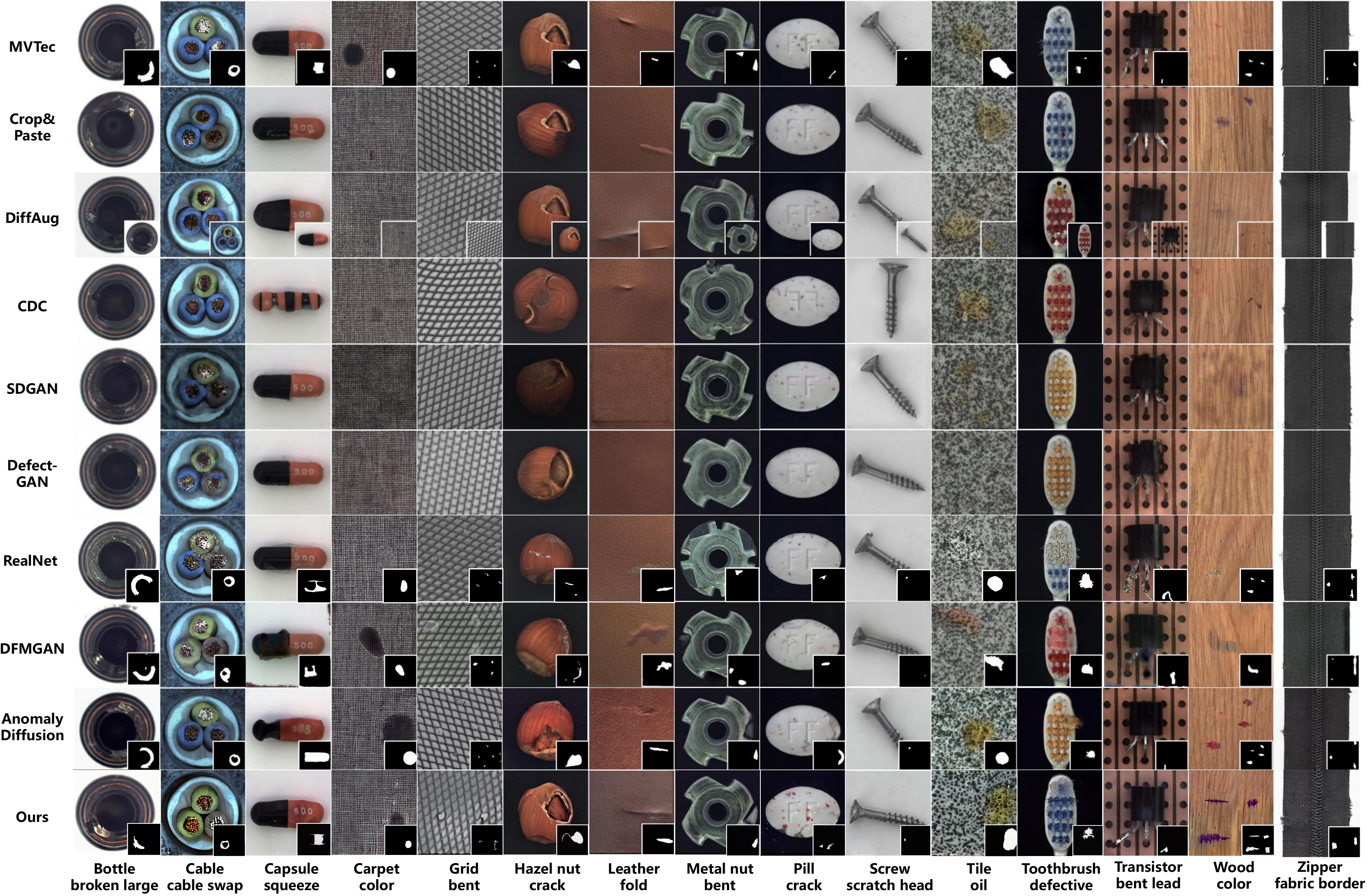}
		
		\caption{Comparison of generation results on MVTec dataset. }
		\label{fig:generation}
	\end{figure*}

	\begin{figure*}[!t]
		\centering
		
		\includegraphics[width=1\linewidth]{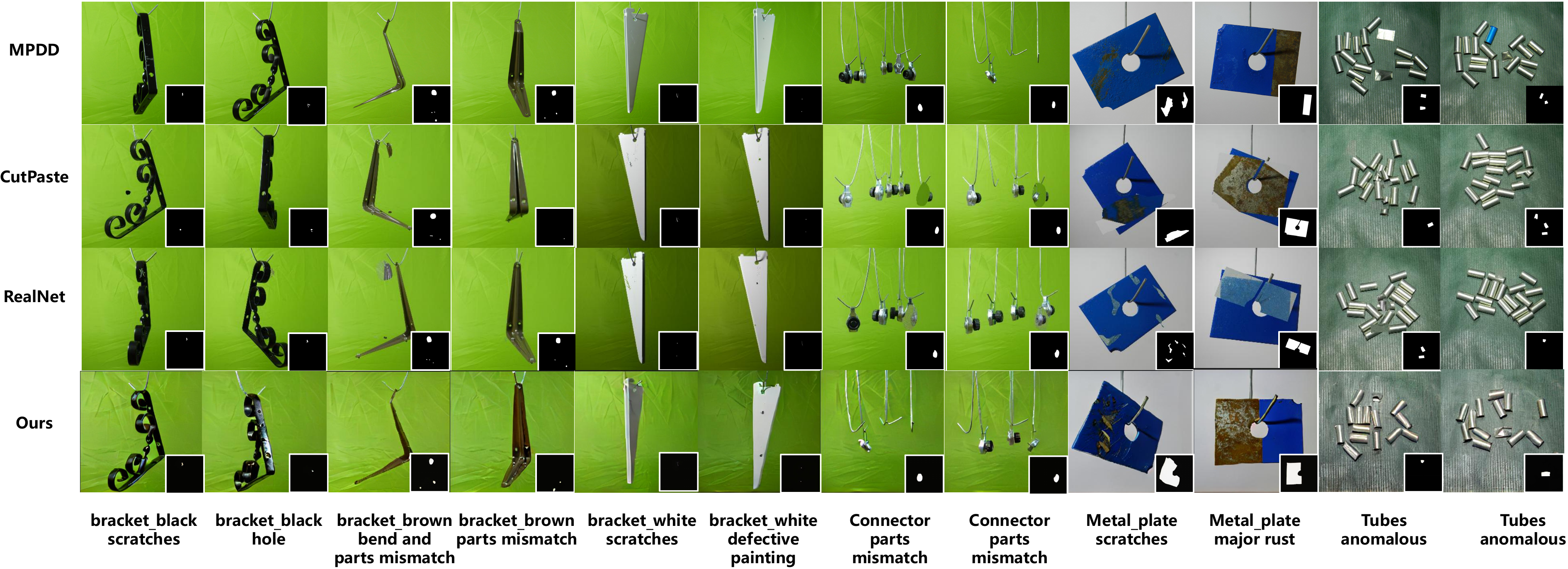}
		
		\caption{Comparison of generation results on MPDD.}
		\label{fig:mpdd_generation}
	\end{figure*}
	\subsection{Training objective}
	The overall training objective can be formulated as:
	\begin{equation}
		\begin{aligned}
			\mathcal{L}_{DH} = E_{\varepsilon(\bx,\by),\br,\epsilon \sim N(0,\boldsymbol{I}),t}||\epsilon_t -\epsilon_{\theta(\bz(t),t,c,\boldsymbol{\mathcal{T}}, \tau_\theta(\br))} ||^2,
		\end{aligned}
		\label{eq: diffusion_loss}
	\end{equation}
	where $\bx,\by$ denotes the input image and annotation, $\bz = \{{\bz^I,\bz^A}\}$ is the concatenated noisy image and noisy annotation feature, $\tau_\theta(\br)$ is the noisy reference image feature encoded with condition encoder $\tau$, $t$ is a random diffusion time step and $\boldsymbol{\mathcal{T}}$ is text embedding.
	
	\begin{table*}[!t]
		\renewcommand\arraystretch{1.2}
		\centering
		\scriptsize
		\caption{Performance Comparison of the different methods with metrics IS and IC-LPIPS on the MVTec Dataset ($\uparrow$ indicates higher is better). The best results are highlighted in bold.}
		\label{tab:generation}
		\setlength{\tabcolsep}{1.95mm}
		\begin{tabular}{p{28pt}c|cc|cc|cc|cc|cc|cc|cc|cc|cc}
			\hline
			\multirow{2}{*}{Category} & & \multicolumn{2}{c}{\underline{DiffAug}\cite{zhao2020differentiable}}  & \multicolumn{2}{c}{\underline{CDC}\cite{ojha2021few}} & \multicolumn{2}{c}{ \underline{Crop$\&$Paste}\cite{lin2021few}}  & \multicolumn{2}{c}{\underline{SDGAN}\cite{niu2020defect}} & \multicolumn{2}{c}{\underline{DefectGAN}\cite{zhang2021defect}}  & \multicolumn{2}{c}{\underline{DFMGAN}\cite{duan2023few}} & \multicolumn{2}{c}{\underline{Ano. diff.}\cite{hu2024anomalydiffusion}} & \multicolumn{2}{c}{\underline{Dual Ano.}\cite{jin2025dual}} & \multicolumn{2}{c}{\underline{Ours}}\\
			& &\multicolumn{2}{c}{IS $\uparrow$ IC-L $\uparrow$}&\multicolumn{2}{c}{IS $\uparrow$ IC-L $\uparrow$}&\multicolumn{2}{c}{IS $\uparrow$ IC-L $\uparrow$}&\multicolumn{2}{c}{IS $\uparrow$ IC-L $\uparrow$}&\multicolumn{2}{c}{IS $\uparrow$ IC-L $\uparrow$}&\multicolumn{2}{c}{IS $\uparrow$ IC-L $\uparrow$}&\multicolumn{2}{c}{IS $\uparrow$ IC-L $\uparrow$}&\multicolumn{2}{c}{IS $\uparrow$ IC-L $\uparrow$}&\multicolumn{2}{c}{IS $\uparrow$ IC-L $\uparrow$}\\
			\midrule
			bottle & & 1.59 & 0.03 & 1.52 & 0.04 & 1.43 & 0.04 & 1.57 & 0.06 & 1.39 & 0.07 & 1.62 & 0.12 & 1.58 & 0.19 & 2.17 & 0.36 & \textbf{2.21} & \textbf{0.38} \\
			cable & & 1.72 & 0.07 & 1.97 & 0.19 & 1.74 & 0.25 & 1.89 & 0.19 & 1.70 & 0.22 & 1.96 & 0.25 & 2.13 & 0.41 & \textbf{2.12} & \textbf{0.43} & 2.07 & 0.35 \\
			capsule & & 1.34 & 0.03 & 1.37 & 0.06 & 1.23 & 0.05 & 1.49 & 0.03 & 1.59 & 0.04 & 1.59 & 0.11 & 1.59 & 0.21 & 1.60 & \textbf{0.31} & \textbf{1.61} & 0.21 \\
			carpet & & 1.19 & 0.06 & 1.25 & 0.03 & 1.17 & 0.11 & 1.18 & 0.11 & 1.24 & 0.12 & 1.23 & 0.13 & 1.16 & 0.24 & 1.36 & \textbf{0.29} & \textbf{1.38} & 0.28 \\
			grid & & 1.96 & 0.06 & 1.97 & 0.07 & 2.00 & 0.12 & 1.95 & 0.10 & 2.01 & 0.12 & 1.97 & 0.13 & 2.04 & 0.44 & \textbf{2.09} & 0.42 & 1.97 & \textbf{0.43} \\
			hazelnut & & 1.67 & 0.05 & 1.97 & 0.05 & 1.74 & 0.21 & 1.85 & 0.16 & 1.87 & 0.19 & 1.93 & 0.24 & 2.13 & 0.31 & 1.91 & 0.35 & \textbf{2.21} & \textbf{0.36} \\
			leather & & 2.07 & 0.06 & 1.80 & 0.07 & 1.47 & 0.14 & 2.04 & 0.12 & 2.12 & 0.14 & 2.06 & 0.17 & 1.94 & 0.41 & 1.88 & 0.34 & \textbf{2.18} & \textbf{0.42} \\
			metal nut & & 1.58 & 0.29 & 1.55 & 0.04 & 1.56 & 0.15 & 1.45 & 0.28 & 1.47 & 0.30 & 1.49 & 0.32 & 1.96 & 0.30 & 1.56 & 0.32 & \textbf{1.94} & \textbf{0.34} \\
			pill & & 1.53 & 0.05 & 1.56 & 0.06 & 1.49 & 0.11 & 1.61 & 0.07 & 1.61 & 0.10 & 1.63 & 0.16 & 1.61 & 0.26 & 1.82 & 0.37 & \textbf{1.85} & \textbf{0.38} \\
			screw & & 1.10 & 0.10 & 1.13 & 0.11 & 1.12 & 0.16 & 1.17 & 0.10 & 1.19 & 0.12 & 1.12 & 0.14 & 1.28 & 0.30 & \textbf{1.34} & \textbf{0.36} & 1.33 & 0.35 \\
			tile & & 1.93 & 0.09 & 2.10 & 0.12 & 1.83 & 0.20 & 2.53 & 0.21 & 2.35 & 0.22 & 2.39 & 0.22 & 2.54 & 0.55 & 2.35 & 0.50 & \textbf{2.62} & \textbf{0.59} \\
			toothbrush & & 1.33 & 0.06 & 1.63 & 0.06 & 1.30 & 0.08 & 1.78 & 0.03 & 1.85 & 0.03 & 1.82 & 0.18 & 1.68 & 0.21 & 2.40 & 0.48 & \textbf{2.42} & \textbf{0.49} \\
			transistor & & 1.34 & 0.05 & 1.61 & 0.13 & 1.39 & 0.15 & 1.76 & 0.13 & 1.47 & 0.13 & 1.64 & 0.25 & 1.57 & 0.34 & 1.69 & 0.33 & \textbf{1.73} & \textbf{0.35} \\
			wood & & 2.05 & 0.30 & 2.05 & 0.03 & 1.95 & 0.23 & 2.12 & 0.25 & 2.19 & 0.29 & 2.12 & 0.35 & 2.33 & 0.37 & 2.21 & 0.40 & \textbf{2.41} & \textbf{0.42} \\
			zipper & & 1.30 & 0.05 & 1.30 & 0.05 & 1.23 & 0.11 & 1.25 & 0.10 & 1.25 & 0.10 & 1.29 & 0.27 & 1.39 & 0.25 & 2.09 & 0.36 & \textbf{2.11} & \textbf{0.38} \\
			\midrule
			Average & & 1.58 & 0.09 & 1.65 & 0.07 & 1.51 & 0.14 & 1.71 & 0.13 & 1.69 & 0.15 & 1.72 & 0.20 & 1.80 & 0.32 & 1.90 & 0.37 & \textbf{1.94} & \textbf{0.39} \\
			\bottomrule
		\end{tabular}
	\end{table*}

	\section{Experiments}
	\label{sec:experiments}
	In this section, we conduct two types of experiments: anomaly generation and anomaly detection. For anomaly generation, we evaluate the diversity and quality of generated samples. For anomaly detection, we train  supervised  detectors leveraging the generated dataset. 
	
	\subsection{Datasets}
	
	\noindent\textbf{ MVTec AD dataset} \cite{bergmann2019mvtec} comprises 5,354 high-resolution color images, including 10 object classes and 5 textures. Image resolutions range from 700$\times$700 to 1,024$\times$1,024, which are standardized to 256$\times$256 for all our experiments. Training samples per class vary from 60 to 320, and the test set includes over 11 anomaly categories like crack, scratch, deformation, hole, color patch, spilled oil and \etc.
	
	\noindent\textbf{MPDD dataset} \cite{jezek2021deep} contains 1,346 images from 6 types
	of industrial metal products with varying lighting conditions, non-uniform backgrounds, and multiple products in
	each image. Furthermore, the placement orientation, shooting distance, and position of the products are also varied.
	
	\noindent\textbf{MVTec LOCO AD} \cite{bergmann2022beyond} dataset includes both structural and logical anomalies. It contains 3,644 images from five different categories. Structural anomalies appear as scratches, dents, or contamination in the manufactured products. Logical anomalies violate constraints, e.g., valid objects in invalid locations or missing required ones.
	
	\noindent\textbf{VisA} \cite{zou2022spot}  comprises 12 distinct object categories, including 9,621 normal samples and 1,200 anomalous samples. The subsets encompass PCBs, capsules, candles, among others. The anomalous samples including surface flaws (e.g., scratches, dents, color spots) and structural anomalies (e.g., misplacements or missing components).

	\subsection{Experimental Setup}
	\noindent\textbf{Implementation Details.}
	For anomaly generation training, we utilize 50\% of the anomalous samples and all normal samples from the MVTec AD, MPDD, MVTec LOCO AD, and VisA datasets. For anomaly detection, we train  supervised detectors leveraging the generated dataset. 
	For network structure, the U-Net encoder comprises 4 down-sample blocks with every block employing both the DDA and SSM modules. The decoder comprises 4 up-sample blocks with every other block employing the DDA and SSM modules. An Adam optimizer is employed with a learning rate of \( \text{lr} = 1 \times 10^{-4} \) and a batch size of 32.

	\noindent\textbf{Evaluation Metrics.}
	For generation, we employed Inception Score (IS) \cite{salimans2016improved} for direct generation quality assessment and Intra-Cluster pairwise LPIPS distance (IC-LPIPS) \cite{ojha2021few} for generation diversity measurement.
	For anomaly inspection, AUROC and Average Precision (AP) were utilized to evaluate the accuracy of anomaly detection and localization.

	\noindent	\textbf{Baseline Methods.} 
	We compared the following state-of-the-art anomaly generation methods: DualAno (CVPR'25),
	Anomaly Diffusion (AAAI'24), RealNet (CVPR'24) , 
	PRN (CVPR'23), DFMGAN (AAAI'23), DualAno, Crop$\&$Paste (ICME'21), CDC (CVPR'21), DefectGAN (WACV'21), DiffAug (NIPS'20), SDGAN (TASE'20).

	\begin{figure*}[!t]
		\centering
		
		\includegraphics[width=0.95\linewidth]{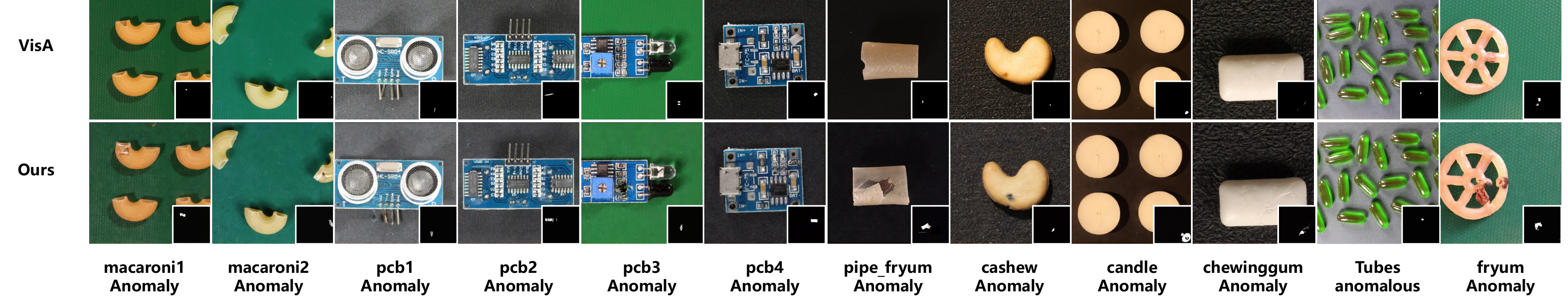}
		
		\caption{Visualization of the generation results obtained by DH-Diff on VisA.}
		\label{fig:Visa}
	\end{figure*}
	\begin{figure*}[!t]
		\centering
		
		\includegraphics[width=0.95\linewidth]{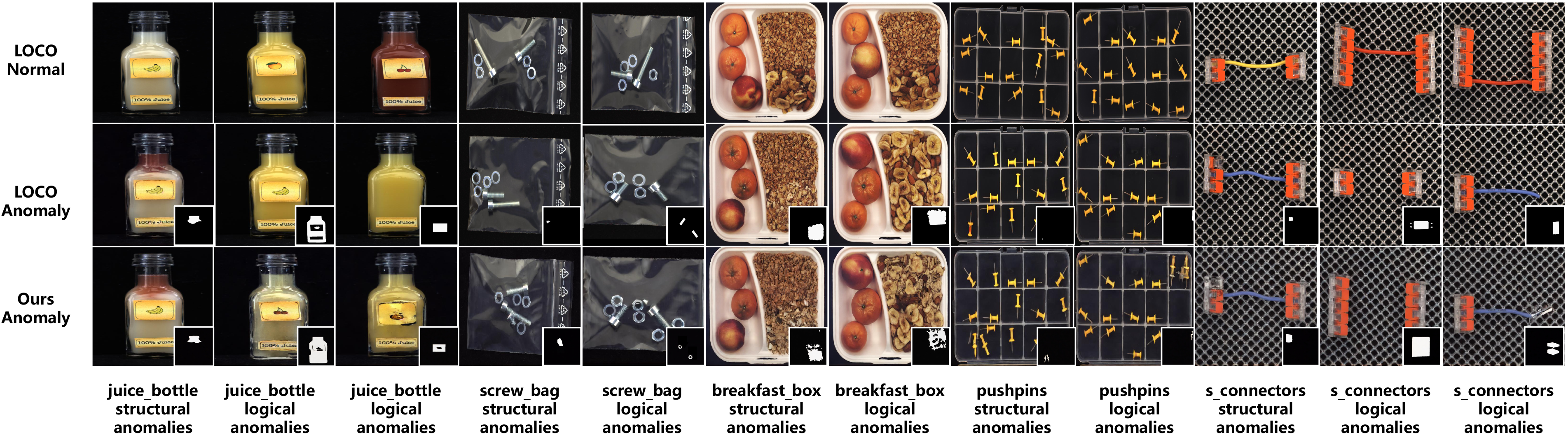}
		
		\caption{Visualization of the generation results obtained by DH-Diff on LOCO.}
		\label{fig:loco}
	\end{figure*}
	
	We classify the compared methods into 2 groups: 1) the models (Crop$\&$Paste\cite{lin2021few},  DualAno\cite{jin2025dual},  
	PRN\cite{zhang2023prototypical},  DFMGAN\cite{duan2023few}), Anomaly Diffusion\cite{hu2024anomalydiffusion} and RealNet\cite{zhang2024realnet}
	that can generate anomalous image-mask pairs, which are
	employed to compare anomaly detection and localization;
	2) the models (DualAno\cite{jin2025dual}, Anomaly Diffusion\cite{hu2024anomalydiffusion}, DiffAug\cite{zhao2020differentiable},  CDC\cite{ojha2021few}, Crop$\&$Paste, SDGAN\cite{niu2020defect},  DefectGAN\cite{zhang2021defect} and DFMGAN\cite{duan2023few}), that can generate specific anomaly types, which are employed to compare
	anomaly generation quality.
	
	\subsection{Comparison in Anomaly Generation}

	\noindent\textbf{Anomaly Generation Quality.} 
	The anomaly generation comparison results is shown in Table \ref{tab:generation}. Due to the random texture cropping used in DRAEM and PRN to simulate anomalies, IC-LPIPS could not be computed for these methods. For each anomaly category, we generated 1,000 anomaly images and annotations to calculate the Inception Score (IS) and IC-LPIPS. The IS criterion evaluates both the authenticity and diversity of the generated images. Specifically, the authenticity score assesses the similarity between the generated images and real images, indicating the likelihood of the generated images being classified as real. The diversity score measures the variation of the generated images across different categories, with higher scores indicating better diversity. The IC-LPIPS metric specifically measures intra-class diversity.
	
	From Table \ref{tab:generation} we can see, DH-Diff achieves the highest IS and IC-LPIPS scores across multiple categories such as bottle, capsule, carpet, hazelnut, leather, pill, tile, wood and \etc. DH-Diff also achieves the highest average score. This demonstrates that the proposed model generates anomaly data with the highest quality and diversity.
	
	Additionally, we show the visual comparisons on the MVTec, MPDD, VisA, and LOCO datasets in Figure 5-8,
	respectively. On MVTec AD dataset, as shown in Figure~\ref{fig:generation},  anomalies generated from the proposed DH-Diff demonstrate superior structural authenticity. For instance, in the `grid bent' category, our generated anomalies exhibit perfectly curved grid structure. In the `capsule squeeze' category, our generated defects are most close to real squeezed capsules. In the `hazelnut crack' category, the generated image features an authentic hazelnut kernel with clear boundaries. Furthermore, the tiny cracks in both the images and masks reflects DH-Diff's fine-grained generation capability. The images generated by DH-Diff possess rich textures with authenticity, and the corresponding masks exhibit a wide diversity.

	\begin{table}[!t]
		\renewcommand\arraystretch{1.1}
		\centering
		\scriptsize
		\setlength{\tabcolsep}{1.5mm}
		\caption{Comparison of DH-Diff and other anomaly synthesis methods on MVTec-AD using Image and Pixel AUROC ($\%$) metrics.}
		\label{tab:local}
		\begin{tabular}{p{30pt}|@{\ \ }c@{\ \ }c@{\ \ }c@{\ \ }c@{\ \ }c}
			\hline
			{Category} & {DRAEM\tiny\cite{zavrtanik2021draem}} & {DFMGAN\tiny\cite{duan2023few}} & {Ano.Diff.\tiny\cite{hu2024anomalydiffusion}} & {RealNet\tiny\cite{zhang2024realnet}} & {Ours} \\
			\midrule
			bottle & (99.3, 96.7) & (99.3, 98.9) & (99.8, 99.4) & (\textbf{100}, 99.3) & (\textbf{100}, \textbf{99.6}) \\
			cable & (72.1, 80.3) & (95.9, 97.2) & (100, 99.2) & (99.2, 98.1) & (\textbf{99.4}, \textbf{98.2}) \\
			capsule & (93.2, 76.2) & (92.8, 79.2) & (99.7, 98.8) & (99.6, 99.3) & (\textbf{99.8}, \textbf{99.1}) \\
			carpet & (95.3, 92.6) & (67.9, 90.6) & (96.7, 98.6) & (\textbf{100}, \textbf{99.3}) & (99.7, 99.4) \\
			grid & (99.8, 99.1) & (73.0, 75.2) & (98.4, 98.3) & (\textbf{100}, \textbf{99.5}) & (99.4, 99.1) \\
			hazelnut & (100, 98.8) & (99.9, 99.7) & (99.8, 99.8) & (\textbf{100}, 99.5) & (\textbf{100}, \textbf{99.8}) \\
			leather & (100, 98.5) & (99.9, 98.5) & (\textbf{100}, 99.8) & (\textbf{100}, 99.8) & (\textbf{100}, \textbf{99.9}) \\
			metal nut & (97.8, 96.9) & (99.3, 99.3) & (\textbf{100}, \textbf{99.8}) & (99.8, 99.6) & (\textbf{100}, 99.7) \\
			pill & (94.4, 95.8) & (68.7, 81.2) & (98, \textbf{99.8}) & (99.1, 99.0) & (\textbf{99.5}, 99.7) \\
			screw & (88.5, 91.0) & (22.3, 58.8) & (96.8, 97.0) & (\textbf{98.9}, \textbf{99.5}) & (98.6, 96.7) \\
			tile & (\textbf{100}, 98.5) & (\textbf{100}, 99.5) & (\textbf{100}, 99.2) & (\textbf{100}, 99.4) & (\textbf{100}, \textbf{99.6}) \\
			toothbrush & (99.4, 93.8) & (\textbf{100}, 96.4) & (\textbf{100}, 99.2) & (99.4, 98.7) & (\textbf{100}, \textbf{99.4}) \\
			transistor & (79.6, 76.5) & (90.8, 96.2) & (\textbf{100}, \textbf{99.2}) & (\textbf{100}, 98) & (\textbf{100}, 99.1) \\
			wood & (\textbf{100}, 98.8) & (98.4, 95.3) & (98.4, \textbf{98.9}) & (99.2, 98.2) & (\textbf{99.5}, 99.1) \\
			zipper & (\textbf{100}, 93.4) & (99.7, 92.9) & (99.9, \textbf{99.4}) & (99.8, 99.2) & (99.6, 99.5) \\
			\midrule
			Average & (94.6, 98.7) & (87.2, 90.9) & (99.2, 99.1) & (\textbf{99.7}, 99.0) & (\textbf{99.7}, \textbf{99.2}) \\
			\bottomrule
		\end{tabular}
	\end{table}

	\begin{table}[!t]
		\renewcommand\arraystretch{1.1}
		\centering
		\scriptsize
		\setlength{\tabcolsep}{1.6mm}
		\caption{Comparison of DH-Diff and other anomaly synthesis methods on MPDD using Image and Pixel AUROC ($\%$) metrics. 
		}
		\label{tab:local-m}
		\begin{tabular}{p{50pt}|cccc}
			\hline
			{Category} & Ours & RealNet\cite{zhang2024realnet} & DTD\cite{cimpoi2014describing}  & CutPaste\cite{li2021cutpaste} \\
			\midrule
			Bracket Black & (\textbf{96.7}, \textbf{99.4})& (94.9, 99.3) & (89.5, 98.9) & (66.4, 96.7) \\
			Bracket Brown & (\textbf{97.2}, \textbf{98.1})& (96.8, 97.8) & (92.9, 97.4) & (95.5, 97.5) \\
			Bracket White & (\textbf{92.1}, \textbf{97.8}) & (88.8, 97.4) & (86.7, 98.6) & (88.4, 96.5) \\
			Connector & (\textbf{100.0}, \textbf{98.9})& (\textbf{100.0}, 97.5) & (99.1, 97.7) & (99.1, 98.5) \\
			Metal Plate & (\textbf{100.0}, \textbf{99.5})& (100.0, 99.3) & (100.0, 99.3) & (99.9, 98.8) \\
			Tubes & (\textbf{98.2}, \textbf{98.6})& (97.5, 97.9) & (92.6, 99.0) & (91.5, 98.1) \\
			\midrule
			AVG & (\textbf{97.2}, \textbf{98.7})& (96.4, 98.2) & (93.5, 98.5) & (90.1, 97.7) \\
			\bottomrule
		\end{tabular}
	\end{table}
	
	In comparison, the Crop$\&$Paste method does not create any new defects. For CDC, the generated data deviates significantly from actual samples on structure and texture. SDGAN and DefectGAN could hardly generate anomalies. DFMGAN and Anomaly diffusion both have visible deviations in terms of authenticity. For example, in the `leather fold' category, the anomaly generated by DFMGAN looks more like liquid marks, while those generated by Anomaly diffusion are more similar to scratches.
	
	The results on MPDD dataset are presented in Figure \ref{fig:mpdd_generation}. The anomalies generated by DH-diff achieve the highest level of structure logicality and generation authenticity, especially in the `Tubes anomalous', `Connector parts mismatch' and `Metal plate major rust' classes. For example, in the `Metal plate major rust' class, our method produces photo-realistic rust appearance, whereas others show evident structure conflicts.

	\begin{table*}[!t]
		\centering
		\setlength{\tabcolsep}{5pt}
		\caption{Comparison on pixel-level anomaly localization (AUROC($\%$),AP($\%$)) on MVTec datasets.}
		\label{tab:comparison}
		\scriptsize
		\begin{tabular}{c|c|c|c|c|c|c|c|c|c|c}
			\hline
			Category  & CFLOW\cite{gudovskiy2022cflow} & SSPCAB\cite{ristea2022self} & CFA\cite{lee2022cfa} & RD4AD\cite{deng2022anomaly} & DevNet\cite{pang2021explainable} & ReDi\cite{xing2025recover} & PRN\cite{zhang2023prototypical} & Ano. Diff.\cite{hu2024anomalydiffusion} & DualAno.\cite{jin2025dual} & \textbf{Ours} \\
			\midrule
			bottle  & (98.8,49.9) & (98.9,88.6) & (98.9,50.9) & (98.8,51.0) & (96.7,67.9) & (98.9,81.5) & (99.4,92.3) & (99.3,94.1) & (99.5,93.4) & (\textbf{99.6},\textbf{95.6}) \\
			cable  & (98.9,72.6) & (93.1,52.1) & (98.4,79.8) & (98.8,77.0) & (97.9,67.6) & (97.9,72.6) & (98.8,78.9) & (\textbf{99.2},\textbf{90.8}) & (97.5,\textbf{82.6}) & (\textbf{98.2},80.8) \\
			capsule  & (99.5,64.0) & (90.4,48.7) & (98.9,71.1) & (99.0,60.5) & (91.1,46.6) & (98.7,42.7) & (98.5,62.2) & (98.8,57.2) & (\textbf{99.5},\textbf{73.2}) & (99.1,71.9) \\
			carpet  & (99.7,67.0) & (92.3,49.1) & (99.1,47.7) & (99.4,46.0) & (94.6,19.6) & (99.2,68.4) & (99.0,82.0) & (98.6,81.2) & (99.4,\textbf{89.1}) & (\textbf{99.5},88.2) \\
			grid  & (99.1,87.8) & (99.6,58.2) & (98.6,82.9) & (98.0,75.4) & (90.2,44.9) & (99.3,50.6) & (98.4,45.7) & (98.3,52.9) & (98.5,57.2) & (\textbf{99.1},\textbf{64.9}) \\
			hazelnut  & (97.9,67.2) & (99.6,94.5) & (98.5,80.2) & (94.2,57.2) & (76.9,46.8) & (99.3,76.4) & (99.7,93.8) & (99.8,96.5) & (99.8,97.7) & (\textbf{99.8},\textbf{98.4}) \\
			leather  & (99.2,91.1) & (97.2,60.3) & (96.2,60.9) & (96.6,53.5) & (94.3,66.2) & (99.5,52.3) & (99.7,69.7) & (99.8,79.6) & (99.9,\textbf{88.8}) & (\textbf{99.9},86.1) \\
			metal nut  & (98.8,78.2) & (99.3,95.1) & (98.6,74.6) & (97.3,53.8) & (93.3,57.4) & (98.0,88.9) & (99.7,98.0) & (99.8,98.7) & (99.6,98.0) & (\textbf{99.7},\textbf{99.2}) \\
			pill  & (98.9,60.3) & (96.5,48.1) & (98.8,67.9) & (98.4,58.1) & (98.9,79.9) & (98.4,79.4) & (99.5,91.3) & (\textbf{99.8},\textbf{97.0}) & (99.6,95.8) & (99.7,94.5) \\
			screw  & (98.8,45.7) & (99.1,62.0) & (98.7,61.4) & (99.1,51.8) & (66.5,21.1) & (99.6,44.8) & (97.5,44.9) & (97.0,51.8) & (\textbf{98.1},\textbf{57.1}) & (96.7,49.3) \\
			tile  & (98.0,86.7) & (99.2,96.3) & (98.6,92.6) & (97.4,78.2) & (88.7,63.9) & (95.7,49.5) & (99.6,96.5) & (99.2,93.9) & (99.7,97.1) & (\textbf{99.8},\textbf{97.7}) \\
			toothbrush  & (99.1,56.9) & (97.5,38.9) & (98.4,61.7) & (99.0,63.1) & (96.3,52.4) & (98.9,62.4) & (99.6,78.1) & (99.1,76.5) & (98.2,68.3) & (\textbf{99.4},\textbf{87.7}) \\
			transistor  & (98.8,40.6) & (85.3,36.5) & (98.6,82.9) & (99.6,50.3) & (55.2,4.4) & (96.1,70.1) & (98.4,85.6) & (\textbf{99.3},\textbf{92.6}) & (98.0,86.7) & (99.1,94.1) \\
			wood  & (98.9,47.2) & (97.2,77.1) & (97.6,25.6) & (99.3,39.1) & (93.1,47.9) & (98.7,55.0) & (97.8,82.6) & (98.9,84.6) & (99.4,91.6) & (\textbf{99.1},\textbf{95.2}) \\
			zipper  & (96.5,63.9) & (98.1,78.2) & (95.9,53.9) & (99.7,52.7) & (92.4,53.1) & (98.9,53.3) & (98.8,77.6) & (99.4,86.0) & (99.6,90.7) & (\textbf{99.5},\textbf{92.1}) \\
			\midrule
			Average  & (98.7,65.3) & (96.2,65.5) & (98.3,66.3) & (98.3,57.8) & (86.4,49.3) & (98.3,63.2) & (99.0,78.6) & (99.1,81.4) & (99.1,84.0) & (\textbf{99.2},\textbf{86.4}) \\
			\bottomrule
		\end{tabular}
	\end{table*}

	\begin{table*}[!t]
		\renewcommand\arraystretch{1.2}
		\centering
		\scriptsize
		\setlength{\tabcolsep}{1.1mm}
		\caption{Comparison of DH-Diff with alternative anomaly detection methods on the MPDD dataset.}
		\label{tab:mpdd_detection}
		\begin{tabular}{p{70pt}|c|c|c|c|c|c|c|c|c|c}
			\hline
			{Metric} & {PatchCore\cite{roth2022towards} } & {CFlow\cite{gudovskiy2022cflow} } & {PaDiM \cite{defard2021padim}} & {SPADE\cite{cohen2020sub} } & {DAGAN\cite{tang2020anomaly} } & {Skip-GANomaly\cite{akccay2019skip}} & {CutPaste\cite{li2021cutpaste}}& {DTD\cite{cimpoi2014describing}} & {RealNet\cite{zhang2024realnet}} & {Ours}  \\
			\midrule
			Image AUROC ($\%$) & 82.1 & 86.1 & 74.8 & 77.1 & 72.5 & 64.8 & 90.1 & 93.5 & 96.3 & \textbf{97.2}\\
			Pixel AUROC ($\%$) & 95.7 & 97.7 & 96.7 & 95.9 & 83.3 & 82.2 & 97.7 & 98.5 & 98.2 & \textbf{98.7}\\
			\bottomrule
		\end{tabular}
	\end{table*}

	\begin{table*}[ht]
		\centering
		\begin{minipage}[t]{0.48\textwidth}
			\centering
			\caption{Comparison on image-level anomaly detection (AUROC($\%$)) on VisA datasets.}
			\label{tab:VisA}
			\scriptsize
			\setlength{\tabcolsep}{3pt}
			\begin{tabular}{l|cccc|c}
				\toprule
				Category& PatchCore\cite{roth2022towards}  & SimpleNet\cite{liu2023simplenet} & DDAD\cite{mousakhan2023anomaly} & GLAD\cite{yao2024glad} & OURS\\
				\midrule
				Candle           & 98.7  & 96.9 & 99.9 & \textbf{99.9}& \textbf{99.9} \\
				Capsules         & 68.8  & 89.5 & \textbf{100}  & 99.1& 99.4 \\
				Cashew           & 97.7  & 94.8 & 94.5 & 98.4& \textbf{98.8} \\
				Chewinggum       & 99.1  & \textbf{100}  & 98.1 & 99.6& 99.7 \\
				Fryum            & 91.6  & 96.6 & 99.0 & 99.4& \textbf{99.6} \\
				Macaroni1        & 90.1  & 97.6 & 99.2 & 99.7& \textbf{99.9} \\
				Macaroni2        & 63.4  & 83.4 & \textbf{99.2} & 98.9& 99.1 \\
				Pcb1             & 96.0  & 99.2 & \textbf{100}  & 99.6& 99.7 \\
				Pcb2             & 95.1  & 99.2 & 99.7 & \textbf{100} & 99.9 \\
				Pcb3             & 93.0  & 98.6 & 97.2 & 99.8& \textbf{99.9} \\
				Pcb4             & \textbf{99.5}  & 98.9 & 100  & 99.9& \textbf{100.0} \\
				Pipe fryum       & 99.0  & 99.2 & \textbf{100}  & 98.9& 99.2 \\
				\textbf{Average} & 91.0  & 96.2 & 98.9 & 99.5& \textbf{99.6} \\
				\bottomrule
			\end{tabular}
			
		\end{minipage}
		\hfill
		\begin{minipage}[t]{0.48\textwidth}
			\centering
			\caption{Comparison on image-level anomaly detection (AUROC($\%$)) on MVTec LOCO datasets.}
			\label{tab:loco}
			\scriptsize
			\setlength{\tabcolsep}{2pt}
			\begin{tabular}{ll|ccc|c}
				\toprule
				Type & Category   & PatchCore\cite{roth2022towards}  & EfficientAD~\cite{batzner2024efficientad} & ComAD~\cite{liu2023component} & OURS \\
				\midrule
				\multirow{5}{*}{LA}
				& Breakfast Box        & 74.8  & 85.5 & 91.1 & \textbf{92.0} \\
				& Juice Bottle         & 93.9  & \textbf{98.4} & 95.0 & 96.2 \\
				& Pushpins             & 63.6  & 97.7 & 95.7 & \textbf{98.0} \\
				& Screw Bag            & 57.8  & 56.7 & 71.9 & \textbf{94.5} \\
				& S. Connectors        & 79.2  & \textbf{95.5} & 93.3 & 94.3 \\
				\midrule
				\multirow{5}{*}{SA}
				& Breakfast Box        & 80.1  & 88.4 & 81.6 & \textbf{94.6} \\
				& Juice Bottle         & 98.5  & 99.7 & 98.2 & \textbf{99.9} \\
				& Pushpins             & 87.9  & 96.1 & 91.1 & \textbf{97.3} \\
				& Screw Bag            & 92.0  & 90.7 & 88.5 & \textbf{94.8} \\
				& S. Connectors        & 88.0  & \textbf{98.5} & 94.9 & 99.0 \\
				\midrule
				\multicolumn{2}{l|}{\textbf{Total Avg}}   & 81.6 & 83.7 & 94.8 & \textbf{95.6} \\
				\bottomrule
			\end{tabular}
			
		\end{minipage}
	\end{table*}

	\subsection{Comparison in Anomaly Detection}
	For further detection evaluation, we first generate 1,000 anomaly images and corresponding annotations for each anomaly category to from the anomaly training set. Then, we leverage the MVTec normal samples as normal training set. The original MVTec test set is employed for test evaluation.

	\noindent\textbf{Anomaly generation for anomaly detection and localization.}  For anomaly detection and localization, we exclusively compare our method with those capable of generating both anomaly images and corresponding annotations, including Anomaly Diffusion, RealNet, Crop$\&$Paste, DRAEM, PRN, and DFMGAN. We employ a U-Net architecture for anomaly localization, and further aggregate the localization results using average pooling to derive confidence scores for image-level anomaly detection, similar to the approach used in DRAEM.
	
	We evaluate the performance using two metrics: Image AUROC (\%) and Pixel AUROC (\%). The results for the MVTec dataset are summarized in Table \ref{tab:local}, while the results for the MPDD dataset are presented in Table \ref{tab:local-m}. Our model consistently outperforms other anomaly generation models across most categories, achieving the highest average scores in both Image AUROC and Pixel AUROC.

	\noindent\textbf{Comparison with alternative anomaly detection models.}
	To rigorously validate the effectiveness of our approach, we conduct comprehensive comparisons against state-of-the-art anomaly detection methods on four different datasets. In the mean time, we adopt the official implementations or publicly available pre-trained models and evaluate all methods on the same test sets to ensure fair comparison.
	
	\textit{MVTec AD Results.} The performance comparison on the MVTec AD dataset, measured by pixel-level AUROC and Average Precision (AP), is summarized in Table~\ref{tab:comparison}. Despite using a simple U-Net backbone, our method achieves the best performance, with an AUROC of $99.2\%$ and an AP of $86.4\%$. These results demonstrate the effectiveness of our synthesized anomaly data in enhancing downstream anomaly localization.
	
	\textit{MPDD Results.} Table~\ref{tab:mpdd_detection} reports the comparison on the MPDD dataset, including methods such as PatchCore~\cite{roth2022towards}, CFLOW~\cite{gudovskiy2022cflow}, PaDiM~\cite{defard2021padim}, SPADE~\cite{cohen2020sub}, DAGAN~\cite{tang2020anomaly}, Skip-GANomaly~\cite{akccay2019skip}, CutPaste~\cite{li2021cutpaste}, DTD~\cite{cimpoi2014describing}, and RealNet~\cite{zhang2024realnet}. Our method achieves the highest performance with an Image-level AUROC of $97.2\%$ and a Pixel-level AUROC of $98.7\%$.
	
	\textit{VisA Results.} Table~\ref{tab:VisA} presents results on the VisA dataset, comparing against PatchCore~\cite{roth2022towards}, SimpleNet~\cite{liu2023simplenet}, DREAM~\cite{zavrtanik2021draem}, DDAD~\cite{mousakhan2023anomaly}, and GLAD~\cite{yao2024glad}. Our model achieves the highest average image-level AUROC of $99.6\%$, further validating the generalizability of our method across diverse industrial scenarios.
	
	\textit{MVTec LOCO Results.} As reported in Table~\ref{tab:loco}, our method achieves competitive performance on the MVTec LOCO dataset, attaining a pixel-level AUROC of $95.6\%$. This result not only highlights the practical utility of our synthesized anomaly data for real-world anomaly localization tasks but also underscores the capability of the diffusion model to accurately generate structural and logical anomaly layouts.

	
	
	
	
	

	\subsection{Ablation Studies}

	\begin{table*}[!t]
		\centering
		\begin{minipage}{0.5\linewidth}
			\centering
			\caption{Impact of different components.}
			
			\begin{tabular}{cccc}
				\hline
				No. & DDA & SSM & AP \\
				\midrule
				(1) & $\times$ & $\times$ & 71.5 \\
				(2) & $\checkmark$ & $\times$ & 78.6 \\
				(3) & $\times$ & $\checkmark$ & 80.3 \\
				(4) & $\checkmark$ & $\checkmark$ & 86.4 \\
				\bottomrule
			\end{tabular}
			\label{tab:ablation_components}
		\end{minipage}%
		\hfill
		\begin{minipage}{0.5\linewidth}
			\centering
			\caption{Impact of network architectures.}
			\begin{tabular}{ccccc|c}
				\hline
				No. & DDA-E & DDA-D & SSM-E & SSM-D & AP \\
				\midrule
				(1) & 4 & $\times$ & 4 & $\times$ & 82.7 \\
				(2) & 2 & $\times$ & 2 & $\times$ & 84.5 \\
				(3) & 2 & 2 & 2 & 2 & 80.6 \\
				(4) & 4 & 4 & 2 & 2 & 86.4 \\
				\bottomrule
			\end{tabular}
			\label{tab:ablation_architecture}
		\end{minipage}
	\end{table*}

	\begin{figure}[!t]
		\centering
		
		\includegraphics[width=0.90\linewidth]{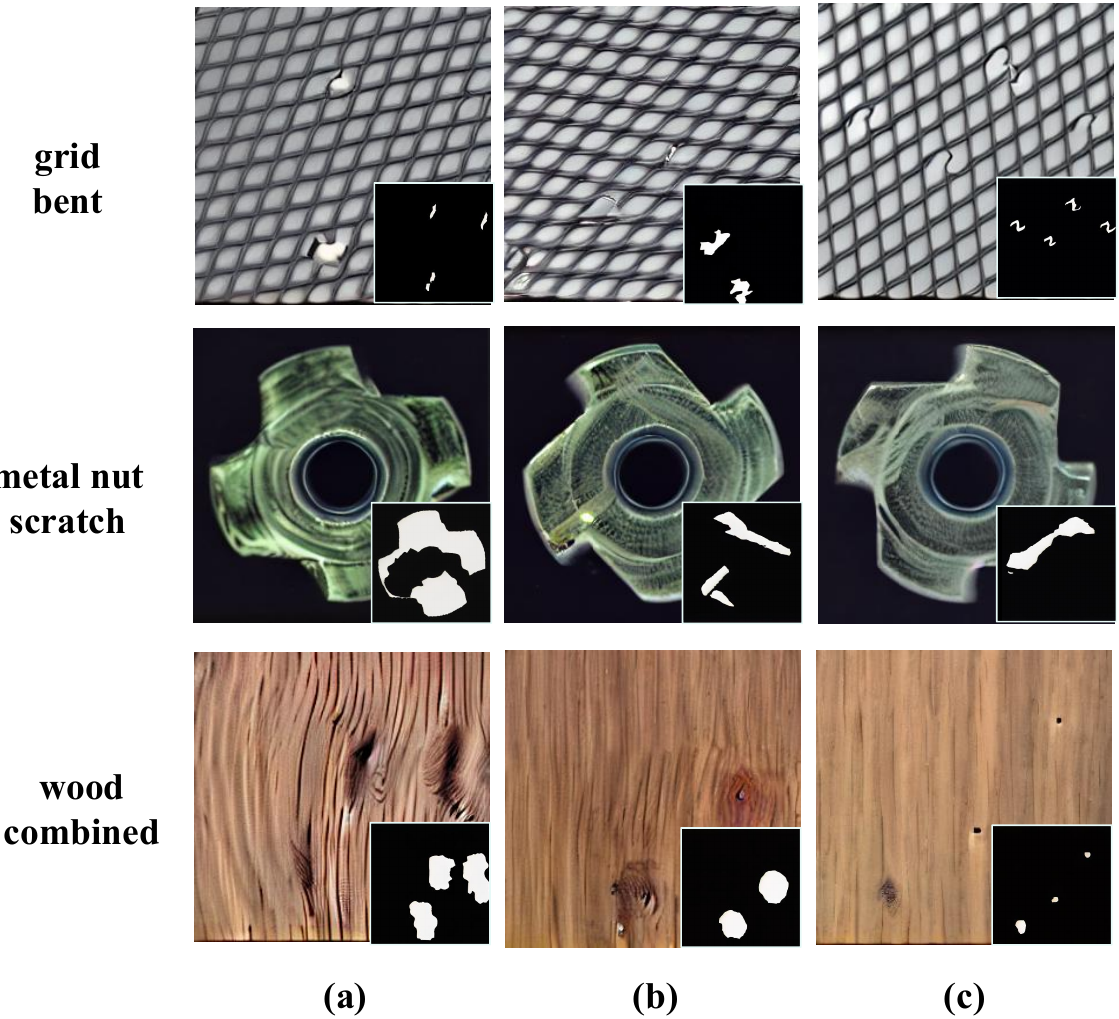}
		
		\caption{Results from ablation studies. (a) Baseline. (b) Baseline + DDA. (c) Baseline + DDA + SSM. Baseline is plain diffusion model.
		}
		\label{fig:ablation}
	\end{figure}
	\noindent\textbf{Impact of different components.}
	We conducted ablation studies on the MVTec AD dataset to explore the effectiveness of each component in our proposed DH-diff model. We chose stable-diffusion [21] as our baseline and gradually added different components in seven experiments: (1) Baseline; (2) Baseline with DDA (decoupled cross-domain attention) added; (3) Adding the SSM (semantic score map modification) module; (4) Adding DDA and SSM. As shown in Table \ref{tab:ablation_components}, our baseline achieved only a 71.5 AUC score. Training with DDA led to an significantly AP improvement of $+7.1\uparrow$ to 78.6. Adding the SSM module to the baseline  increased model performance by $+8.8\uparrow$ . Adding both DDA and SSM modules to the baseline significantly improved anomaly detection performance to 86.4. The result strongly demonstrate the effectiveness of proposed DDA and SSM.

	Furthermore, we provide visual demonstrations of the impact of different components in our model, as shown in Figure \ref{fig:ablation}. The baseline model shows chaos in both image features and annotation boundaries. After adding DDA, the generated anomaly images exhibit a significant improvement in authenticity. However, there are still structure conflicts in categories like `wood combined', where the lines are mismatched, and inconsistencies between the generated anomaly and the annotation, such as in the `metal nut scratch' category, where the annotation includes background areas. After incorporating SSM, our model achieves the highest generation authenticity and more reasonable structure consistency.

	\noindent\textbf{Impact of network architecture.}
	To comprehensively assess the influence of network architecture, particularly the number and placement of the proposed modules, we conducted a series of ablation studies on the MVTec AD dataset. The experimental settings are summarized in Table~\ref{tab:ablation_architecture}: 
	1) inserting DDA and SSM into every block of the encoder;
	2) inserting DDA and SSM into every other block of the encoder;
	3) inserting DDA and SSM into every other block of both the encoder and decoder;
	4) inserting DDA and SSM into every block of the encoder and every other block of the encoder.
	
	The results demonstrate that the optimal performance is obtained when DDA and SSM are applied to all encoder blocks, while only every other block in decoder. This configuration effectively balances domain separation and semantic alignment, and is therefore adopted as the final model design in our framework.


	\section{Conclusion}
	\label{sec:conclusion}
	This paper proposed an innovative cross-domain one backbone diffusion framework for simultaneously anomaly image and annotation generation.  Specifically, to address feature entanglement problem, we proposed the DDA (Decoupled Domain Attention) module, which divides input into the image domain and annotation domain, and reconstructs them separately. To resolve structure conflicts, we introduced SSM (score map modification), which achieves reasonable and identical defect region generation through semantic heat map alignment. Extensive experiments showed DH-diff significantly outperforms state-of-the-art methods in generation authenticity and diversity, enhancing downstream anomaly detection tasks.
	
	%
	%
	%
	%
	%
	%
	%
	%
	%
	%

	\small
	\bibliographystyle{ieee_fullname}
	\bibliography{egbib}

	\begin{IEEEbiography}
		[{\includegraphics[width=1in,clip,keepaspectratio]{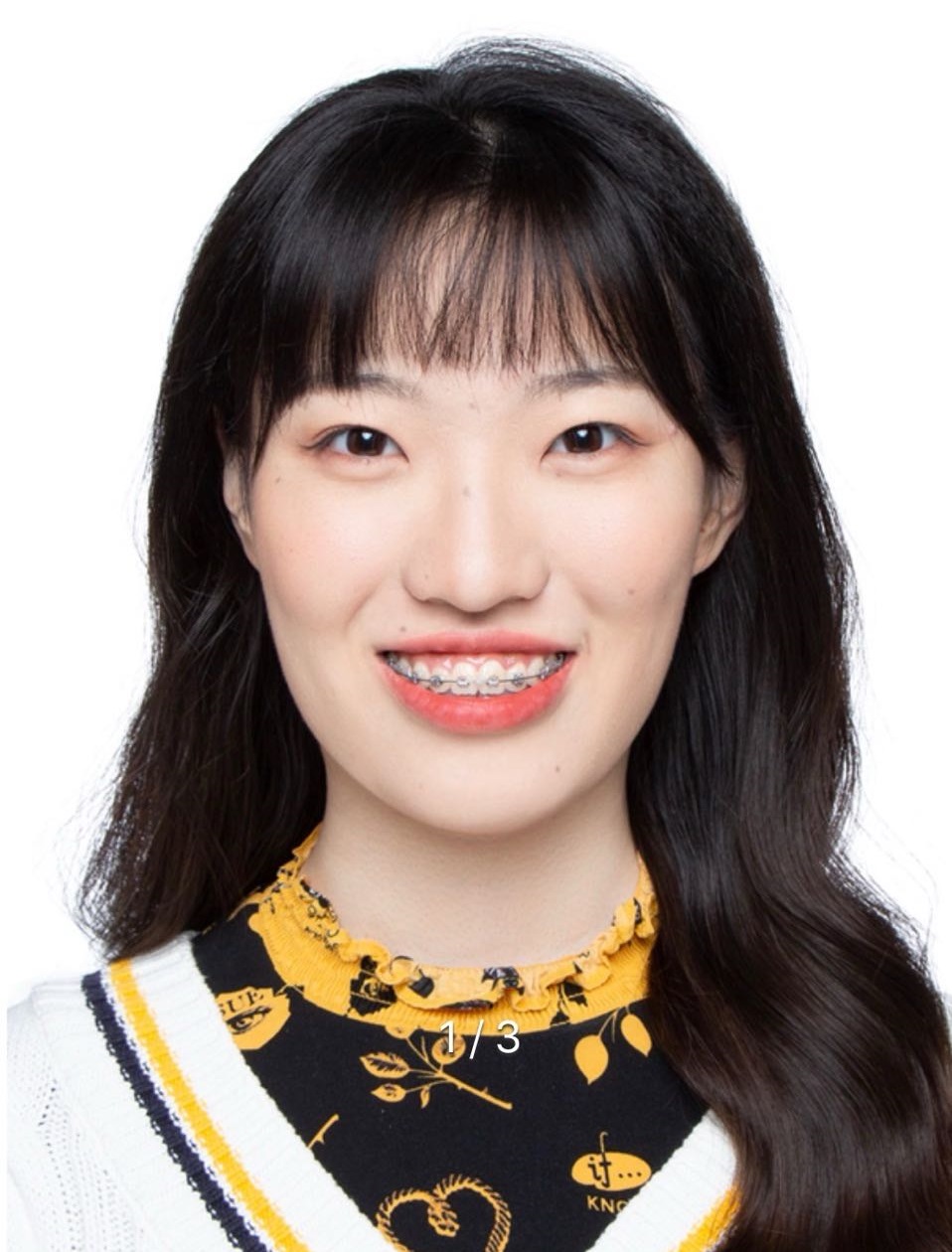}}]
		{Linchun Wu} received the B.S. degree in Automation from Wuhan University in 2019. She is currently working towards the Ph.D. Degree in Computer Science at Wuhan University, China. Her research interests include image and video generation, industrial anomaly detection.
	\end{IEEEbiography}
	
	\begin{IEEEbiography}
		[{\includegraphics[width=1in,clip,keepaspectratio]{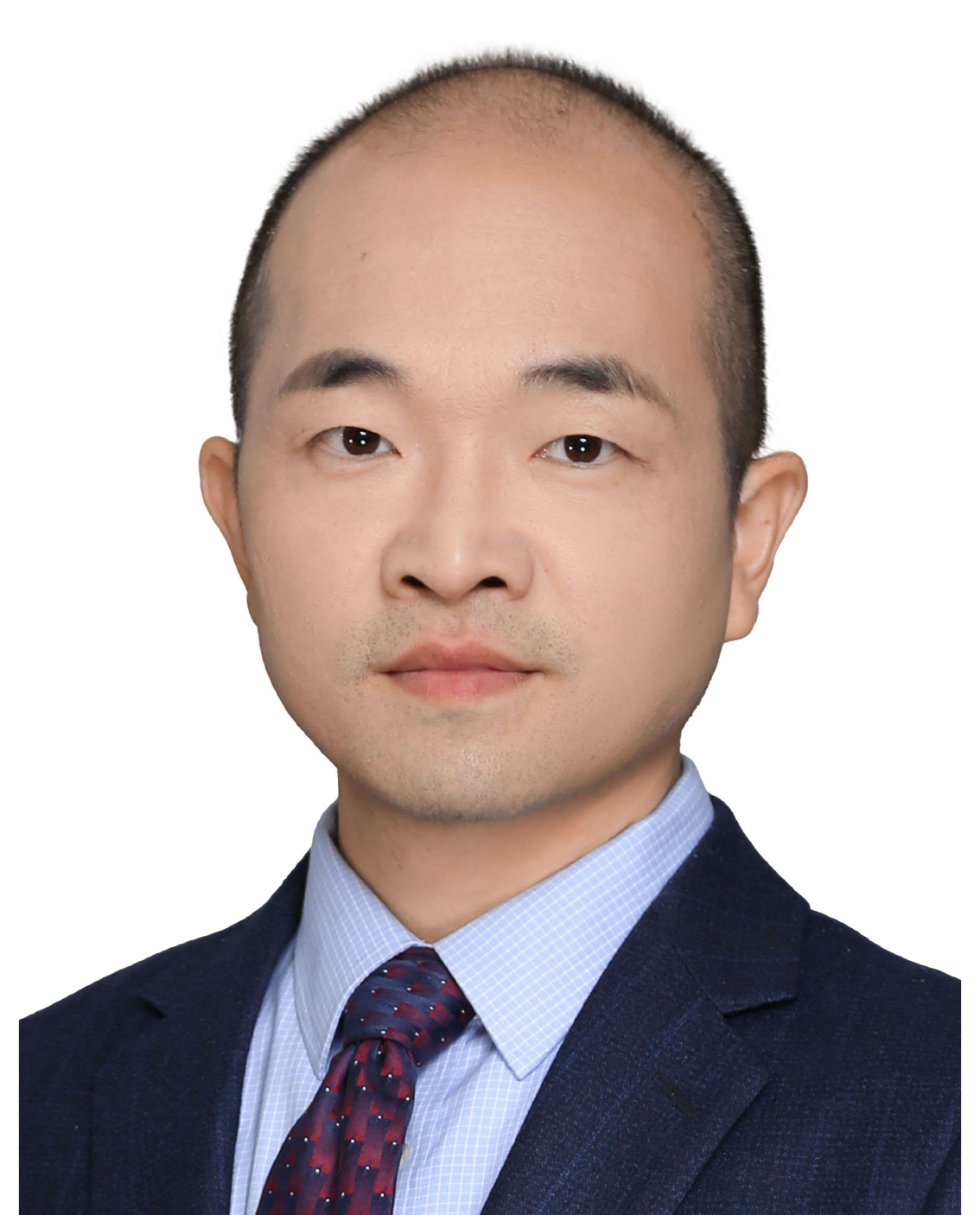}}]
		{Qin Zou} (Senior Member, IEEE) received the B.E. degree in information engineering and the Ph.D. degree in computer vision both from Wuhan University, China. From 2010 to 2011, he was a visiting Ph.D. student at the Computer Vision Laboratory, University of South Carolina, USA. He has been affiliated with the School of Computer Science at Wuhan University since 2012 and currently serves as a Full Professor. His research interests include machine vision, machine learning, and robotics. Since 2023, he has served as an Associate Editor for Signal, Image and Video Processing. He was a co-recipient of the National Technology Invention Award of China in 2015.
	\end{IEEEbiography}
	
	\begin{IEEEbiography}
		[{\includegraphics[width=1in,clip,keepaspectratio]{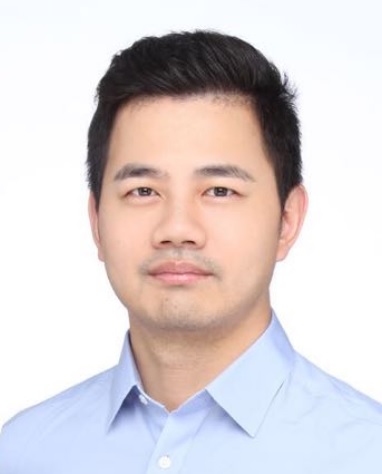}}]
		{Xianbiao Qi} received his B.E. degree in 2008 and a Ph.D. degree in 2015 from Beijing University of Posts and Telecommunications (BUPT). From Jan 2014 to Nov 2015, he was a researcher in the Center of Machine Vision group at Oulu University of Finland. He was a postdoctoral researcher at Hong Kong Polytechnic University (PolyU) from May 2016 to May 2018. He was a senior image expert at Ping An Property \& Casualty Insurance company from April 2019 to July 2021. From August 2021 to March 2024, he was a senior research scientist in International Digital Economy Academy (IDEA), Shenzhen, China. Currently, he is the Chief AI expert in Shenzhen Intellifusion Technologies Co Ltd. His research interests lie in optimization of deep learning, large-scale image and multimodal pretrained model and computer vision.
	\end{IEEEbiography}

	\begin{IEEEbiography}
		[{\includegraphics[width=1in,clip,keepaspectratio]{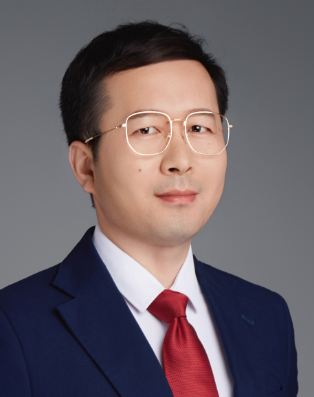}}]
		{Bo Du} received the Ph.D. degree from Wuhan University, China. He is currently a Luojia Distinguished Professor at Wuhan University, where he serves as the Dean of the School of Computer Science, the Executive Vice Dean of the Artificial Intelligence Research Institute, the Director of the National Engineering Research Center for Multimedia Software, and the Director of the Hubei Key Laboratory of Multimedia and Network Communication Engineering. He is a recipient of the National Science Fund for Distinguished Young Scholars (2022).  His research interests include artificial intelligence, computer vision, pattern recognition, and data mining. He has presided over or participated in more than 30 national and provincial-level research projects, published 23 ESI highly cited or hot papers, with over 8000 SCI citations, authored 3 books, and holds 35 granted national invention patents.
		
	\end{IEEEbiography}
	
	\begin{IEEEbiography}
		[{\includegraphics[width=1in,clip,keepaspectratio]{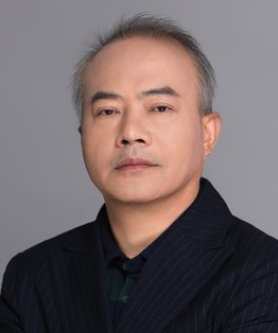}}]
		{Zhongyuan Wang} (M’13) received the Ph.D. degree in communication and information systems from Wuhan University, Wuhan, China. He is currently a Professor with the School of Computer Science, Wuhan University. He has authored or coauthored more than 100 scientific papers, including TPAMI, TIP, CVPR, ICCV, etc. His research interests include biometrics, computer vision and multimedia.
	\end{IEEEbiography}
	
	\begin{IEEEbiography}
		[{\includegraphics[width=1in,clip,keepaspectratio]{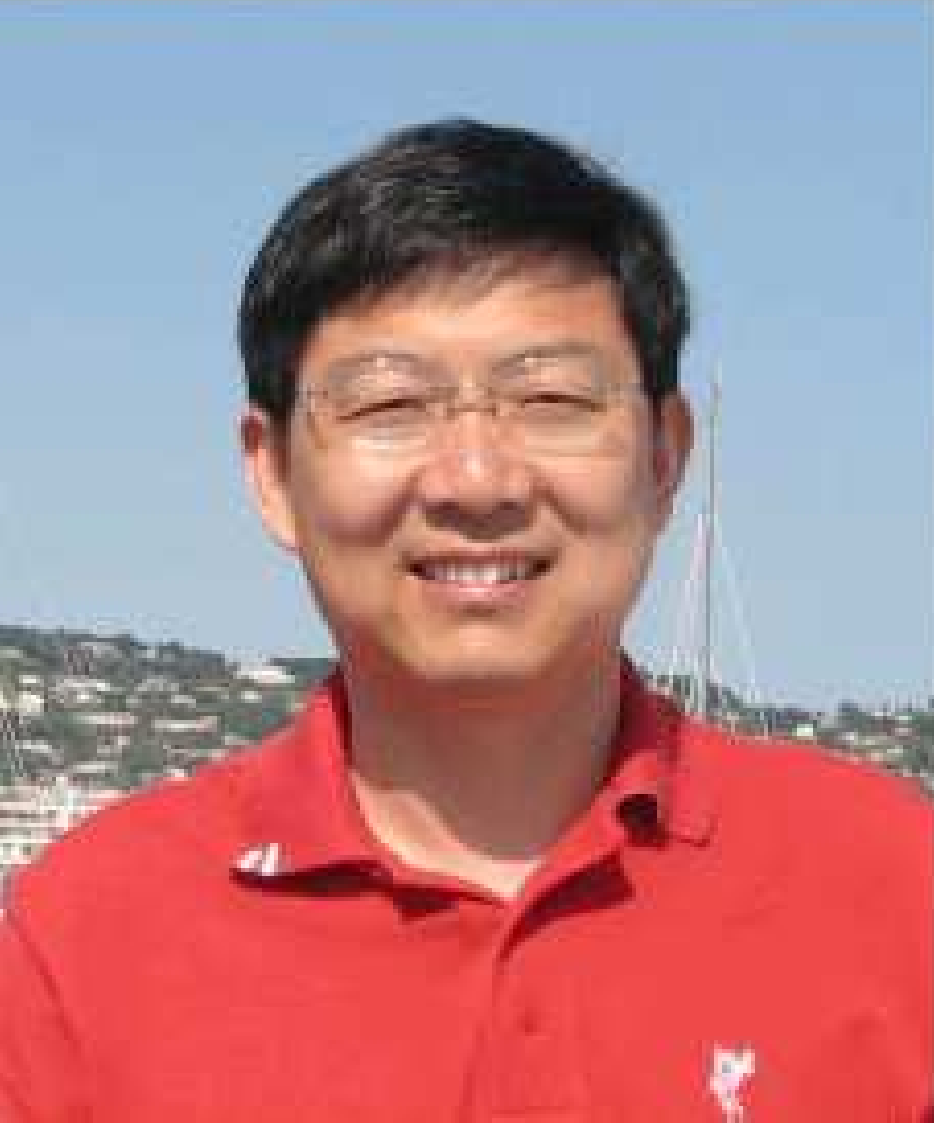}}]
		{Qingquan Li} received his PhD degree from Wuhan  Technical University of Surveying and Mapping, China. He is a professor of Shenzhen University, China; a professor of the State Key Laboratory of Information Engineering in Surveying, Mapping and Remote Sensing, Wuhan University; He is an academician of Chinese Academy of Engineering; He previously served as the Chief Scientist for the 973 Program and as a member of the expert group for the Ministry of Science and Technology's 863 Program. He has long been dedicated to theoretical innovation and equipment development in dynamic precision engineering measurement. The research outcomes have been extensively applied in the national infrastructure inspection and monitoring. He has been awarded the National Technology Invention Prize. His research areas include  dynamic precision engineering measurement, industrial surveying and mapping, and infrastructure defect inspection.
	\end{IEEEbiography}

\end{document}